\documentclass[smallcondensed]{svjour3}     
\smartqed  
\usepackage{hyperref}
\usepackage{url}
\usepackage{graphicx} 
\usepackage{subfigure}
\usepackage{natbib} 
\usepackage{algorithm} 
\usepackage{algorithmic} 
\usepackage{amsmath,amssymb}
\usepackage{wrapfig}
\usepackage{comment}
\usepackage{color}
\newtheorem{Definition}{{\bf Definition}}
\newtheorem{Proposition}{{\bf Proposition}}
\newtheorem{Theorem}{{\bf Theorem}}

\newtheorem{Example}{{\bf Example}}

\begin{document}

\title{Model-based Kernel Sum Rule: Kernel Bayesian Inference with Probabilistic Models
\thanks{This research was partly supported by JSPS KAKENHI (B) 22300098, MEXT Grant-in-Aid for
Scientific Research on Innovative Areas 25120012, JSPS Wakate (B) 26870821 and the ERC action StG 757275 / PANAMA.
}
}
\subtitle{}


\author{Yu Nishiyama         \and
        Motonobu Kanagawa    \and
        Arthur Gretton    \and
        Kenji Fukumizu    
}


\institute{Yu Nishiyama \at
              The University of Electro-Communications, Japan \\
              \email{ynishiyam@gmail.com}           
           \and
           Motonobu Kanagawa \at
             University of T\"ubingen, Germany \\
              \email{motonobu.kanagawa@uni-tuebingen.de} 
           \and
           Arthur Gretton \at
              Gatsby Computational Neuroscience Unit, University College London, England \\
              \email{arthur.gretton@gmail.com} 
           \and
           Kenji Fukumizu \at
              The Institute of Statistical Mathematics, Japan \\
              \email{fukumizu@ism.ac.jp} 
}

\date{Received: date / Accepted: date}

\maketitle

\begin{abstract}

Kernel Bayesian inference is a principled approach to nonparametric inference in probabilistic graphical models, where probabilistic relationships between variables are learned from data in a nonparametric manner.
Various algorithms of kernel Bayesian inference have been developed by combining kernelized basic probabilistic operations such as the kernel sum rule and kernel Bayes' rule. 
However, the current framework is fully nonparametric, and
it does not allow a user to flexibly combine nonparametric and model-based inferences.
This is inefficient when there are good probabilistic models (or simulation models) available for some parts of a graphical model; this is in particular true in scientific fields where ``models'' are the central topic of study.
Our contribution in this paper is to introduce a novel approach, termed the {\em model-based kernel sum rule} (Mb-KSR), to combine a probabilistic model and kernel Bayesian inference.
By combining the Mb-KSR with the existing kernelized probabilistic rules, one can develop various algorithms for hybrid (i.e., nonparametric and model-based) inferences.
As an illustrative example, we consider Bayesian filtering in a state space model, where typically there exists an accurate probabilistic model for the state transition process.
We propose a novel filtering method that combines model-based inference for the state transition process and data-driven, nonparametric inference for the observation generating process.
We empirically validate our approach with synthetic and real-data experiments, the latter being the problem of vision-based mobile robot localization in robotics, which illustrates the effectiveness of the proposed hybrid approach.


\end{abstract}

\section{Introduction} \label{sec:Introduction}

{Kernel mean embedding of distributions} \citep{Smola_etal_ALT2007,KernelEmbeddingofConditionalDistributions,KernelMeanEmbeddingofDistributions_review2017} is a framework for representing, comparing and estimating probability distributions using positive definite kernels and the Reproducing Kernel Hilbert Spaces (RKHS).
In this framework, all distributions are represented as corresponding elements, called {\em kernel means}, in an RKHS, and comparison and estimation of distributions are carried out by comparison and estimation of the corresponding kernel means.
The Maximum Mean Discrepancy  \citep{DBLP:journals/jmlr/GrettonBRSS12} and the Hilbert-Schmidt Independence Criterion  \citep{GreBouSmoSch05} are representative examples of approaches based on comparison of kernel means; the former is a distance between probability distributions and the latter is a measure of dependence between random variables, both enjoying empirical successes and being widely employed in the machine learning literature \citep[Chapter 3]{KernelMeanEmbeddingofDistributions_review2017}.

{\em Kernel Bayesian inference} \citep{DBLP:journals/jmlr/SongGBLG11,KernelEmbeddingofConditionalDistributions,KernelBayes'Rule_BayesianInferencewithPositiveDefiniteKernels} is a nonparametric approach to Bayesian inference based on estimation of kernel means.
In this approach, statistical relationships between any two random variables, say $X \in \mathcal{X}$ and $Z \in \mathcal{Z}$ with $\mathcal{X}$ and $\mathcal{Z}$ being measurable spaces, are nonparametrically learnt from training data consisting of pairs $(X_1,Z_1),\dots,(X_n,Z_n) \in \mathcal{X} \times \mathcal{Z}$ of instances.
The approach is useful when the relationship between $X$ and $Z$ is complicated and thus it is difficult to design an appropriate parametric model for the relationship; it is effective when the modeller instead has good knowledge about similarities between objects in each domain, expressed as similarity functions or kernels of the form $k_\mathcal{X}(x,x')$ and $k_\mathcal{Z}(z,z')$.
For instance, the relationship can be complicated when the structures of the two domains $\mathcal{X}$ and $\mathcal{Z}$ are very different, e.g., $\mathcal{X}$ may be a three dimensional space describing locations, $\mathcal{Z}$ may be a space of images, and the relationship between $X \in \mathcal{X}$ and $Z \in \mathcal{Z}$ is such that $Z$ is a vision image taken at a location $X$; since such images are highly dependent on the environment, it is not straightforward to provide a model description for that relationship.
In this specific example, however, one can define appropriate similarity functions or kernels; the Euclidean distance may provide a good similarity measure for locations, and there are also a number of kernels for images developed in computer vision (e.g., \citealt{Schmid06beyondbags}).
Given a sufficient number of training examples and appropriate kernels, kernel Bayesian inference enables an algorithm to learn such complicated relationships in a nonparametric manner, often with strong theoretical guarantees \citep{CapDev07,DBLP:journals/corr/abs-1205-4656,KernelBayes'Rule_BayesianInferencewithPositiveDefiniteKernels}.

As standard Bayesian inference consists of basic probabilistic rules such as the {\em sum rule}, {\em chain rule} and {\em Bayes' rule}, kernel Bayesian inference consists of kernelized probabilistic rules such as the {\em kernel sum rule}, {\em kernel chain rule} and {\em kernel Bayes' rule} \citep{KernelEmbeddingofConditionalDistributions}.
By combining these kernelized rules, one can develop {\em fully-nonparametric} methods for various inference problems in probabilistic graphical models, where probabilistic relationships between any two random variables are learnt nonparametrically from training data, as described above.
Examples include methods for filtering and smoothing in state space models \citep{KernelBayes'Rule_BayesianInferencewithPositiveDefiniteKernels,KernelBayesSmoothingAISTATS2016,KernelMonteCarlo2016}, belief propagation in pairwise Markov random fields \citep{DBLP:journals/jmlr/SongGBLG11}, likelihood-free inference for simulator-based statistical models \citep{nakagome2013kernel,pmlr-v48-mitrovic16,KajYamKanFuk18,HsuRam19}, and reinforcement learning or control problems \citep{DBLP:journals/corr/abs-1206-4655,DBLP:conf/uai/NishiyamaBGF12,PathIntegralControlbyReproducingKernelHilbertSpaceEmbedding2013,HilbertSpaceEmbeddingsofPredictiveStateRepresentations,MorMarRam18}.
We refer to \citet[Chapter 4]{KernelMeanEmbeddingofDistributions_review2017} for a survey of further applications. 
Typical advantages of the approaches based on kernel Bayesian inference are that i) they are equipped with theoretical convergence guarantees; ii) they are less prone to suffer from the curse of dimensionality, when compared to traditional nonparametric methods such as those based on kernel density estimation\footnote{Note that kernel density estimation is a classical nonparametric approach studied in the statistics literature, where ``kernels'' refer to smoothing kernels, but not reproducing kernels in general. One should not confuse this classical approach with kernel mean embeddings, which is rather a new framework for statistical inference developed in the last decade.} \citep{Sil86}; and iii) they may be applied to non-standard spaces of structured data such as graphs, strings, images and texts, by using appropriate kernels designed for such structured data \citep{LearningwithKernels2002}.

We argue, however, that the fully-nonparametric nature is both an advantage and a {\em limitation} of the current framework of kernel Bayesian inference.
It is an advantage when there is no part of a graphical model for which a good probabilistic model exists, while it becomes a limitation when there does {\em exist} a good model for some part of the graphical model.
Even in the latter case, kernel Bayesian inference requires a user to prepare training data for that part and an algorithm to learn the probabilistic relationship nonparametrically; this is inefficient, given that there already exists a probabilistic model.
The contribution of this paper is to propose an approach to making direct use of a probabilistic model in kernel Bayesian inference, when it is available.
Before describing this, we first explain below why and when such an approach can be useful.

\subsection{Combining Probabilistic Models and Kernel Bayesian Inference} \label{sec:intro-combination}

\begin{figure}[t]
\begin{center}
\includegraphics[width = 0.8\linewidth]{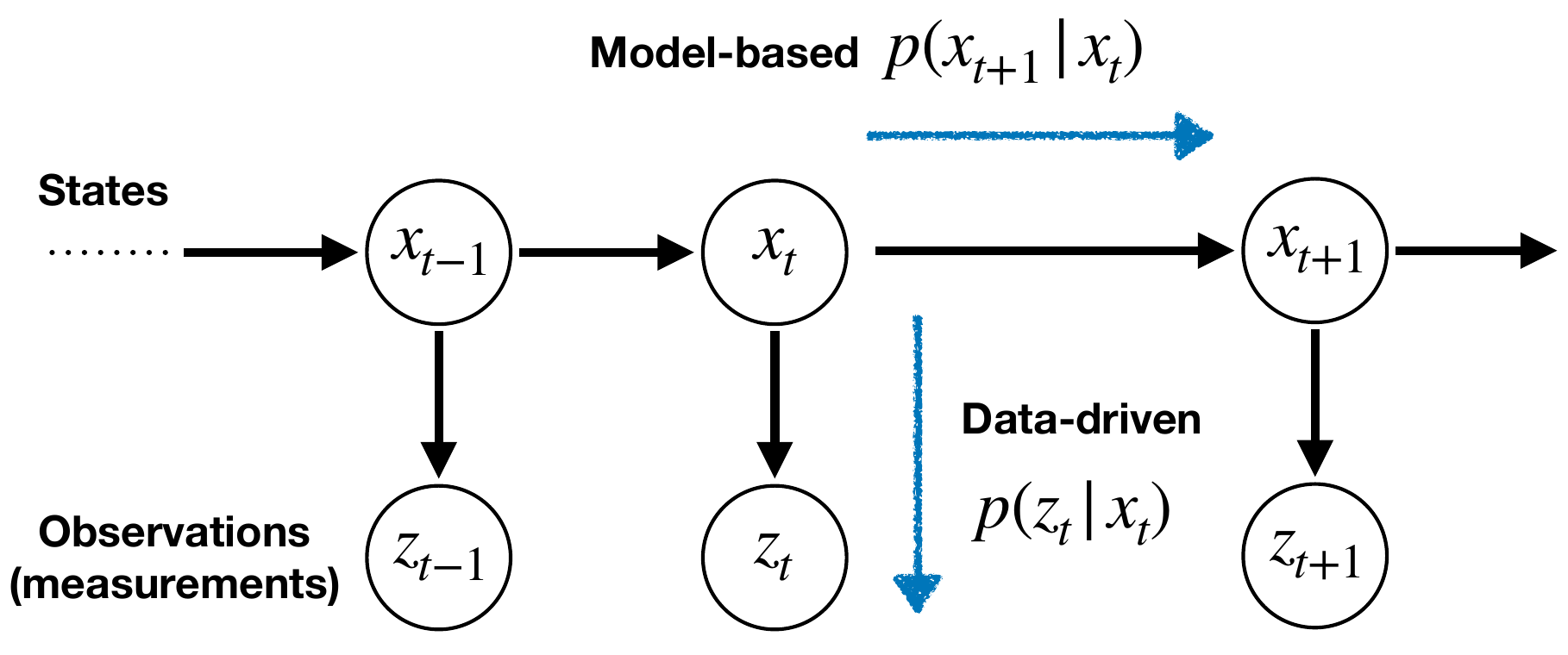}
\caption{A graphical description of a state space model, where $x_t$ represent states and $z_t$ observations (or measurements). 
In this paper we consider a situation where there exists a good probabilistic model for the state-transition probability $p(x_{t+1}|x_t)$, while the observation process $p(z_t | x_t)$ is complicated and to be dealt with in a data-driven, nonparametric way.}
\label{fig:state-space-model}
\end{center}
\end{figure}

An illustrative example is given by the task of {\em filtering} in {\em state space models}; see Fig.~\ref{fig:state-space-model} for a graphical model.
A state space model consists of two kinds of variables: {\em states} $x_1,\dots,x_t,\dots,x_T$, which are the unknown quantities of interest, and {\em observations} $z_1,\dots,z_t,\dots,z_T$, which are measurements regarding the states. 
Here discrete time intervals are considered, and $t = 1,\dots,T$ denote time indices with $T$ being the number of time steps.
The states evolve according to a Markov process determined by the {\em state transition model} $p(x_{t+1}|x_t)$ describing the conditional probability of the next state $x_{t+1}$ given the current one $x_t$.
The observation $z_t$ at time $t$ is generated depending only on the corresponding state $x_t$ following the {\em observation model}, the conditional probability of $z_t$ given $x_t$.
The task of filtering is to provide a (probabilistic) estimate of the state $x_t$ at each time $t$ using the observations $z_1,\dots,z_t$ provided up to that time; this is to be done sequentially for every time step $t = 1,\dots,T$.

In various scientific fields that study time-evolving phenomena such as climate science, social science, econometrics and epidemiology, one of the main problems is {\em prediction} (or {\em forecasting}) of unseen quantities of interest that will realize in the future. 
Formulated within a state space model, such quantities of interest are defined as states $x_1,x_2,\dots,x_T$ of the system. 
Given an estimate of the initial state $x_1$, predictions of the states $x_2,\dots,x_T$ in the future are to be made on the basis of the transition model $p(x_{t+1}|x_t)$, often performed in the form of computer simulation.
A problem of such predictions is, however, that errors (which may be stochastic and/or numerical) accumulate over time, and predictions of the states increasingly become unreliable.
To mitigate this issue, one needs to make corrections to predictions on the basis of  available observations $z_1, z_2, \dots, z_T$ about the states; such procedure is known as {\em data assimilation} in the literature, and formulated as filtering in the state space model \citep{Eve09}.

When solving the filtering problem with kernel Bayesian inference, one needs to express each of the transition model $p(x_{t+1}|x_t)$ and the observation model $p(z_t|x_t)$ by training data: one needs to prepare examples of state-observation pairs $(X_i,Z_i)_{i=1}^n$ for the observation model, and transition examples $(\tilde{X}_i, \tilde{X}'_i)_{i=1}^m$ for the transition model, where $\tilde{X}_i$ denotes a state at a certain time and  $\tilde{X}'_i$ the subsequent state \citep{Song+al:icml2010hilbert,KernelBayes'Rule_BayesianInferencewithPositiveDefiniteKernels}.
However, when there already exists a good probabilistic model for state transitions, it is not efficient to re-express the model by examples and learn it nonparametrically.
This is indeed the case in the scientific fields mentioned above, where a central topic of study is to provide an accurate but succinct model description for the evolution of the states $x_1,x_2,\dots,x_T$, which may take the form of (ordinary or partial) differential equations or that of multi-agent systems \citep{Winsberg2010}.
Therefore it is desirable to make kernel Bayesian inference being able to directly make use of an available transition model in filtering.

\subsection{Contributions}
Our contribution is to propose a simple yet novel approach to combining the nonparametric methodology of kernel Bayesian inference and model-based inference with probabilistic models.
A key ingredient of Bayesian inference in general is the sum rule, i.e., marginalization or integration of variables, which is used for propagating probabilities in graphical models.
The proposed approach, termed {\em Model-based Kernel Sum Rule (Mb-KSR)}, realizes the sum rule in the framework of kernel Bayesian inference, directly making use of an available probabilistic model.
(To avoid confusion, we henceforth refer to the kernel sum rule proposed by \cite{Song+al:icml2010hilbert} as the {\em Nonparametric  Kernel Sum Rule (NP-KSR)}.)
It is based on analytic representations of conditional kernel mean embeddings \citep{KernelEmbeddingofConditionalDistributions}, employing a kernel that is compatible with the probabilistic model under consideration.
For instance, the use of a Gaussian kernel enables the MB-KSR if the probabilistic model is an additive Gaussian noise model. 
A richer framework of hybrid (i.e., nonparametric and model-based) kernel Bayesian inference can be obtained by combining the Mb-KSR with existing kernelized probabilistic rules such as the NP-KSR, kernel chain rule and kernel Bayes' rule.

As an illustrative example, we propose a novel method for filtering in a state space model, under the setting discussed in Sect.~\ref{sec:intro-combination} (see Fig.~\ref{fig:state-space-model}).
The proposed algorithm is based on hybrid kernel Bayesian inference, realized as a combination of the Mb-KSR and the kernel Bayes' rule. 
It directly makes use of a transition model $p(x_{t+1}|x_{t})$ via the Mb-KSR, while utilizing training data consisting of state-observation pairs $(X_1,Z_1), \dots, (X_n,Z_n)$ to learn the observation model nonparametrically.
Thus it is useful in prediction or forecasting applications where the relationship between observations and states is not easy to be modeled, but examples can be given for it; an example from robotics is given below.
This method has an advantage over the fully-nonparametric filtering method based on kernel Bayesian inference \citep{KernelBayes'Rule_BayesianInferencewithPositiveDefiniteKernels} as it makes use of the transition model $p(x_{t+1}|x_t)$ in a direct manner, without re-expressing it by state transition examples and learning it nonparametrically.
This advantage is more significant when the transition model $p(x_{t+1}|x_t)$ is time-dependent (i.e., it is not invariant over time); for instance this is the case when the transition model involves control signals, as for the case in robotics.

One illustrative application of our filtering method is mobile robot localization in robotics, which we deal with in Sect.~\ref{sec:Experiments}. 
In this problem, there is a robot moving in a certain environment such as a building.
The task is to sequentially estimate the positions of the robot as it moves, using measurements obtained from sensors of the robot such as vision images and signal strengths.
Thus, formulated as a state space model, the state $x_t$ is the position of the robot, and the observation $z_t$ is the sensor information.
The transition model $p(x_{t+1}|x_t)$ describes how the robot's position changes in a short time; since this follows a mechanical law, there is a good probabilistic model such as an odometry motion model \citep[Sect.~2.3.2]{ProbabilisticRobotics2005}.
On the other hand, the observation model $p(z_t | x_t)$ is hard to provide a model description, since the sensor information $z_t$ are highly dependent on the environment and can be noisy; e.g., it may depend on the arrangement of rooms and be affected by people walking in the building.  
Nevertheless, one can make use of position-sensor examples $(X_1,Z_1),\dots,(X_n,Z_n)$ collected before the test phase using an expensive radar system or by manual annotation \citep{pronobis2009ijrr}.

The remainder of this paper is organized as follows.
We briefly discuss related work in Sect.~\ref{sec:related} and review the framework of kernel Bayesian inference in Sect. \ref{sec:Preliminaries:NonparametricKernelBayesianInference}.
We propose the Mb-KSR in Sect.~\ref{sec:KernelBayesianInferencewithProbabilisticModels}, providing also a theoretical guarantee for it, as manifested in Proposition \ref{Prop:consistencyMb-KSR}.
We then develop the filtering algorithm in Sect. \ref{sec:FilteringinStateSpaceModels}.
Numerical experiments to validate the effectiveness of the proposed approach are reported in Sect. \ref{sec:Experiments}.
For simplicity of presentation, we only focus on the Mb-KSR combined with additive Gaussian noise models in this paper, but our framework also allows for other noise models, as described in Appendix \ref{sec:ConditionalKernelMeanofProbabilisticModel}.

\section{Related Work} \label{sec:related}



We review here existing methods for filtering in state space methods that are related to our filtering method proposed in Sect.~\ref{sec:FilteringinStateSpaceModels}.
For related work on kernel Bayesian inference, we refer to Sect.~\ref{sec:Introduction} and \ref{sec:Preliminaries:NonparametricKernelBayesianInference}.
\begin{itemize}
\item The Kalman filters \citep{Kal60,JulUhl04} and particle methods \citep{Doucet2001,DouJoh11} are standard approaches to filtering in state space models. 
These methods typically assume that the domains of states and observations are subsets of Euclidean spaces, and require probabilistic models for both the state transition and observation processes be defined. 
On the other hand, the proposed filtering method does not assume a probabilistic model for the observation process, and can learn it nonparametrically from training data, even when the domain of observations is a non-Euclidean space.

\item \citet{GP-BayesFilters_Bayesian_filtering_using_Gaussian_process_prediction_Ko2009, Deisenroth_analyticmoment-based,Robust_Filtering_and_Smoothing_with_Gaussian_Processes_Deisenroth2012} proposed methods for nonparametric filtering and smoothing in state space models based on Gaussian processes (GPs).
Their methods nonparametrically learn both the state transition model and the observation model using Gaussian process regression \citep{GaussianProcessesforMachineLearning}, assuming training data are available for the two models. 
A method based on kernel Bayesian inference has been shown to achieve superior performance compared to GP-based methods, in particular when the Gaussian noise assumption by the GP-approaches is not satisfied (e.g., when noises are multi-modal)  \citep{McCalman2013, Function_Embeddings_for_Multi-modal_Bayesian_Inference_McCalman2013}.

\item Nonparametric belief propagation \citep{NonparametricBeliefPropagationACM2010}, which deals with generic graphical models, nonparametrically estimates the probability density functions of messages and marginals using kernel density estimation (KDE)  \citep{Sil86}. 
In contrast, in kernel Bayesian inference density functions themselves are not estimated, but rather their kernel mean embeddings in an RKHS are learned from data.
\citet{DBLP:journals/jmlr/SongGBLG11} proposed a belief propagation algorithm based on kernel Bayesian inference, which outperforms nonparametric belief propagation.

\item The filtering method proposed by \citet[Sect. 4.3]{KernelBayes'Rule_BayesianInferencewithPositiveDefiniteKernels} is fully nonparametric: It nonparametrically learns both the observation process and the state transition process from training data on the basis of kernel Bayesian inference. 
On the other hand, the proposed filtering method combines model-based inference for the state transition process using an available probabilistic model, and nonparametric kernel Bayesian inference for the observation process.

\item The kernel Monte Carlo filter \citep{KernelMonteCarlo2016} combines nonparametric kernel Bayesian inference with a sampling method. 
The algorithm generates Monte Carlo samples from a probabilistic model for the state transition process based, and estimates the kernel means of forward probabilities based on them.
In contrast, the proposed filtering method does not use sampling but utilizes the analytic expressions of the kernel means of probabilistic models.\footnote{Intuitively, the relationship between the kernel Monte Carlo filter and the proposed filter may be understood as something similar to the relationship between a particle filter and a Kalman filter: As the Kalman filter does not require sampling and makes use of the analytic solutions of required integrals, the proposed filter does not perform sampling and uses analytic solutions of the integrals required for computing kernel means.}
\end{itemize}

\section{Preliminaries: Nonparametric Kernel Bayesian Inference}
\label{sec:Preliminaries:NonparametricKernelBayesianInference} 
In this section we briefly review the framework of kernel Bayesian inference.
We begin by reviewing basic properties of positive definite kernels and reproducing kernel Hilbert spaces (RKHS) in Sect.~\ref{sec:preliminary-kernel-RKHS}, and those of kernel mean embeddings in Sect.~\ref{sec:kernel-mean-embedding} and \ref{sec:kernel-mean-estimator}; we refer to \citet[Sect.~4]{SupportVectorMachines2008} for details of the former, and to \citet[Sect.~3]{KernelMeanEmbeddingofDistributions_review2017} for those of the latter.
We then describe basics of kernel Bayesian inference in Sect.~\ref{sec:conditional-mean-embeddings}, \ref{sec:NP-KSR} and \ref{sec:KBR}; further details including various applications can be found in \citet{KernelEmbeddingofConditionalDistributions} and
\citet[Sect.~4]{KernelMeanEmbeddingofDistributions_review2017}.


\subsection{Positive Definite Kernels and Reproducing Kernel Hilbert Space (RKHS)}
\label{sec:preliminary-kernel-RKHS}
We first introduce positive definite kernels and RKHSs.
Let $\mathcal{X}$ be an arbitrary nonempty set. 
A symmetric function $k\colon \mathcal{X}\times \mathcal{X} \rightarrow \mathbb{R}$ is called a {\em positive definite kernel} if it satisfies the following: $\forall n \in \mathbb{N}$ and $\forall x_1,\ldots, x_n \in \mathcal{X}$, the matrix $G \in \mathbb{R}^{n \times n}$ with elements $G_{i,j} = k(x_i,x_j)$ is positive semidefinite. 
Such a matrix $G$ is referred to as a Gram matrix. 
For simplicity we may refer to a positive definite kernel $k$ just as a {\em kernel} in this paper.
For instance, kernels on $\mathcal{X} = \mathbb{R}^m$ include the Gaussian kernel $k(x,x') = \exp(- \| x - x' \|^2 / \gamma^2)$ and the Laplace kernel $k(x,x') = \exp(- \| x - x' \| / \gamma)$, where $\gamma > 0$.

For each fixed $x \in \mathcal{X}$, $k(\cdot, x)$ denotes a function of the first argument: $x' \to k(x',x)$ for $x' \in \mathcal{X}$. 
A kernel $k$ is called {\em bounded} if $\sup_{x \in \mathcal{X}} k(x,x) < \infty$.  
When $\mathcal{X} = \mathbb{R}^m$, a kernel $k$ called {\it shift invariant} if there exists a function $\kappa\colon \mathbb{R}^{m} \rightarrow \mathbb{R}$ such that $k(x,x')=\kappa(x-x')$, $\forall x,x' \in \mathbb{R}^{m}$. 
For instance, Gaussian, Laplace, Mat\`ern and inverse (multi-)quadratic kernels are shift-invariant kernels; see \citet[Sect.~4.2]{GaussianProcessesforMachineLearning}.


Let $\mathcal{H}$ be a Hilbert space consisting of functions on $\mathcal{X}$, with ${\left\langle \cdot ,\cdot \right\rangle _{\mathcal{H}}}$ being its inner product.
The space $\mathcal{H}$ is called a {\em Reproducing Kernel Hilbert Space (RKHS)}, if there exists a positive definite kernel $k:\mathcal{X} \times \mathcal{X}$ satisfying the following two properties:
\begin{eqnarray}
&& k(\cdot,x) \in \mathcal{H}, \quad \forall x \in \mathcal{X}, \nonumber \\
&& f(x) = {\left\langle {f,k(\cdot,x)} \right\rangle _{\mathcal{H}}}, \quad \forall f \in \mathcal{H},\ \forall x \in \mathcal{X}  \label{eq:reproducingproperty},
\end{eqnarray}
where \eqref{eq:reproducingproperty} is called the {\em reproducing property}; thus $k$ is called the {\em reproducing kernel} of the RKHS $\mathcal{H}$.

Conversely, for {\em any} positive definite kernel $k$, there exists a uniquely associated RKHS $\mathcal{H}$ for which $k$ is the reproducing kernel; this fact is known as the {\em Moore-Aronszajn theorem} \citep{Aronszajn1950}. 
Using the kernel $k$, the associate RKHS $\mathcal{H}$ can be written as the closure of the linear span of functions $k(\cdot,x)$:
$$
\mathcal{H} = \overline{ {\rm span}\left\{ k(\cdot,x):\ x \in \mathcal{X} \right\} }.
$$


\subsection{Kernel Mean Embeddings of Distributions}
\label{sec:kernel-mean-embedding}
We introduce the concept of kernel mean embeddings of distributions, a framework for representing, comparing and estimating probability distributions using kernels and RKHSs.
To this end, let $\mathcal{X}$ be a measurable space and $\mathcal{M}_1(\mathcal{X})$ be the set of all probability distributions on $\mathcal{X}$.
Let $k$ be a measurable kernel on $\mathcal{X}$ and $\mathcal{H}$ be the associated RKHS.
For any probability distribution $P \in \mathcal{M}_1(\mathcal{X})$, we define its representation in $\mathcal{H}$ as an element called the {\it kernel mean}, defined as the Bochner integral of $k(\cdot,x) \in \mathcal{H}$ with respect to $P$:
\begin{eqnarray}
{m_P} := \int k(\cdot, x) dP(x) \in \mathcal{H}. \label{eq:MeanEmbedding}
\end{eqnarray}
If $k$ is bounded, then the kernel mean \eqref{eq:MeanEmbedding} is well-defined and exists for all $P \in \mathcal{M}_1(\mathcal{X})$ \citep[Lemma 3.1]{KernelMeanEmbeddingofDistributions_review2017}.
Throughout this paper, we thus assume that kernels are bounded. 
Being an element in $\mathcal{H}$, the kernel mean $m_P$ itself is a function such that $m_P(x') = \int k(x',x)dP(x)$ for $x' \in \mathcal{X}$.


The definition \eqref{eq:MeanEmbedding} induces a mapping (or embedding; thus the approach is called kernel mean {\em embedding}) from the set of probability distributions  $\mathcal{M}_1(\mathcal{X})$ to the RKHS $\mathcal{H}$: $P \in \mathcal{M}_1(\mathcal{X}) \to m_P \in \mathcal{H}$.
If this mapping is one-to-one, that is $m_P = m_Q$ holds if and only if $P = Q$ for $P,Q \in \mathcal{M}_1(\mathcal{X})$, then the reproducing kernel $k$ of $\mathcal{H}$ is called {\em characteristic} \citep{DBLP:journals/jmlr/FukumizuBJ03,Sriperumbudur_JMLR2010,Simon-Gabriel-JMLR2018}. 
For example, frequently used kernels on $\mathbb{R}^m$ such as Gaussian, Mat\'ern and Laplace kernels are characteristic; see, e.g., \citet{Sriperumbudur_JMLR2010,CharacteristicKernelsandInfinitelyDivisibleDistributions} for other examples. 
If $k$ is characteristic, then any $P \in \mathcal{M}_1(\mathcal{X})$ is uniquely associated with its kernel mean $m_P$; in other words, $m_P$ uniquely identifies the embedded distribution $P$, and thus $m_P$ contains all information about $P$.
Therefore, when required to estimate certain properties of $P$ from data, one can instead focus on estimation of its kernel mean $m_P$; this is discussed in Sect.~\ref{sec:kernel-mean-estimator} below.

An important property regarding the kernel mean \eqref{eq:MeanEmbedding} is that it is the representer of integrals with respect to $P$ in $\mathcal{H}$: for any $f \in \mathcal{H}$, it holds that

\begin{eqnarray}
{\left\langle {{m_P},f} \right\rangle _{\mathcal{H}}} = \left< \int k(\cdot,x)dP(x), f\right>_{\mathcal{H}}    = \int \left<  k(\cdot,x), f\right>_{\mathcal{H}} dP(x) = \int f(x) dP(x), \label{eq:expectationproperty}
\end{eqnarray}

where the last equality follows from the reproducing property \eqref{eq:reproducingproperty}.
Another important property is that it induces a distance or a metric on the set of probability distributions $\mathcal{M}_1(\mathcal{X})$:
A distance between two distributions $P, Q \in \mathcal{M}_1(\mathcal{X})$ is defined as the RKHS distance between their kernel means $m_P, m_Q \in \mathcal{H}$:
\begin{eqnarray*}
\left\| m_P-m_Q \right\|_{{ \mathcal{H}}} = \sup_{ \| f \|_{\mathcal{H}} \le 1 }   \int f(x)dP(x) -  \int f(x)dQ(x)  ,
\end{eqnarray*}
where the expression in the right side is known as the {\em Maximum Mean Discrepancy (MMD)}; see    \citet[Lemma 4]{DBLP:journals/jmlr/GrettonBRSS12} for a proof of the above identity.
MMD is an instance of integral probability metrics, and its relationships to other metrics such as the Wasserstein distance have been studied in the literature \citep{On_the_empirical_estimation_of_integral_probability_metrics,Simon-Gabriel-JMLR2018}.



\subsection{Empirical Estimation of Kernel Means}
\label{sec:kernel-mean-estimator}
In Bayesian inference, one is required to estimate or approximate a certain probability distribution $P$ (or its density function) from data, where $P$ may be a posterior distribution or a predictive distribution of certain quantities of interest. 
In kernel Bayesian inference, one instead estimates its kernel mean $m_P$ from data; this is justified as long as the kernel $k$ is characteristic.

We explain here how one can estimate a kernel mean in general.
Assume that one is interested in estimation of the kernel mean $m_P$ \eqref{eq:MeanEmbedding}.
In general, an estimator of $m_P$ takes the form of a weighted sum 
\begin{equation}
\hat{m}_{P}=\sum_{i=1}^{n} w_{i} k(\cdot, X_i) \label{eq:nonparametric_kernel_mean_estimator},
\end{equation}   
where $w_1,\dots,w_n \in \mathbb{R}$ are some weights (some of which can be negative) and $X_1,\dots,X_n \in \mathcal{X}$ are some points.
For instance, assume that one is given i.i.d.~sample points $X_1,\dots,X_n$ from $P$; then the equal weights $w_1 = \cdots w_n = 1/n$ make \eqref{eq:nonparametric_kernel_mean_estimator} is a consistent estimator with convergence rate $\| {m}_{P} -\hat{m}_{P}  \| _{\mathcal{H}} = O_{p}(n^{- \frac{1}{2}})$ \citep{Smola_etal_ALT2007,ThoSriMua17}. In the setting of Bayesian inference, on the other hand, i.i.d.~sample points from the target distribution $P$ are not provided, and thus
 $X_1,\dots,X_n$ in \eqref{eq:nonparametric_kernel_mean_estimator} cannot be i.i.d.~with $P$. 
Therefore the weights $w_1,\dots,w_n$ need to be calculated in an appropriate way depending on the target $P$ and available data; we will see concrete examples in Sect.~\ref{sec:conditional-mean-embeddings}, \ref{sec:NP-KSR} and \ref{sec:KBR} below.


From (\ref{eq:expectationproperty}), the kernel mean estimate (\ref{eq:nonparametric_kernel_mean_estimator}) can be used to estimate the integral $\int f(x)dP(x)$ of any $f \in \mathcal{H}$ with respect to $P$ as a weighted sum of function values:
\begin{eqnarray}
\int f(x)dP(x) = \left< m_P, f \right>_{\mathcal{H}} \approx {\left\langle {{\hat m_P},f} \right\rangle _{\mathcal{H}}} = \sum_{i=1}^{n} w_{i} f(X_i),  \label{eq:expectationestimate}
\end{eqnarray}
where the last expression follows from the reproducing property \eqref{eq:reproducingproperty}.
In fact, by the Cauchy-Schwartz inequality, it can be shown that $\left| \int f(x)dP(x) - \sum_{i=1}^{n} w_{i} f(X_i)  \right| \leq \| f \|_{\mathcal{H}} \| \hat m_{P}- m_{P} \|_{\mathcal{H}}$.
Therefore, if $\hat{m}_{P}$ is a consistent estimator of $m_P$ such that $|| \hat m_{P}- m_{P} ||_{\mathcal{H}} \mathop \to \limits 0$ as $n \to \infty$, then the weighted sum in (\ref{eq:expectationestimate}) is also consistent in the sense that $\left| \int f(x)dP(x) - \sum_{i=1}^{n} w_{i} f(X_i)  \right| \to 0$ as $n \to \infty$. 
The consistency and convergence rates in the case where $f$ does not belong to $\mathcal{H}$ have also been studied \citep{KanagawaNIPS2016,Kanagawa2019}.

\subsection{Conditional Kernel Mean Embeddings}
\label{sec:conditional-mean-embeddings}

For simplicity of presentation, we henceforth assume that probability distributions under consideration have {\em density functions} with some reference measures; this applies to the rest of this paper.
However we emphasize that this assumption is generally not necessary both in practice and theory. 
This can be seen from how the estimators below are constructed, and from theoretical results in the literature. 

We first describe a kernel mean estimator of the form \eqref{eq:nonparametric_kernel_mean_estimator} when $P$ is a {\em conditional} distribution \citep{Song+al:icml2010hilbert}.
To describe this, let $\mathcal{X}$ and $\mathcal{Y}$ be two measurable spaces, and let $p(y|x)$ be a conditional density function of $y \in \mathcal{Y}$ given $x \in \mathcal{X}$.
Define a kernel $k_\mathcal{X}$ on $\mathcal{X}$ and let $\mathcal{H}_\mathcal{X}$ be the associated RKHS.
Similarly, let $k_\mathcal{Y}$ be a kernel on $\mathcal{Y}$ and $\mathcal{H}_{\mathcal{Y}}$ be its RKHS.

Assume that $p(y|x)$ is unknown, but training data $\{ (X_i, Y_i)\}_{i=1}^{n} \subset \mathcal{X} \times \mathcal{Y}$ approximating it are available; usually they are assumed to be i.i.d.~with a joint probability  $p(x,y) = p(y|x)p(x)$, where $p(x)$ is some density function on $\mathcal{X}$.
Using the training data $\{ (X_i, Y_i)\}_{i=1}^{n}$, we are interested in estimating the kernel mean of the conditional probability $p(y|x)$ on $\mathcal{Y}$ for a given $x$:
\begin{eqnarray}
m_{\mathcal{Y}| x}:=   \int k_\mathcal{Y}(\cdot,y) p(y|x) dy \in \mathcal{H_Y}, \label{eq:conditionalkernelmeandef}
\end{eqnarray}
which we call the {\em conditional kernel mean}.

\citet{Song+al:icml2010hilbert} proposed the following estimator of \eqref{eq:conditionalkernelmeandef}:
\begin{eqnarray}
 && \hat m_{\mathcal{Y}| x} = \sum_{j=1}^{n} w_{j}(x) k_{\mathcal{Y}}(\cdot, Y_j), \label{eq:conditionalkernelmean} \\
 && w(x):= (w_1(x),\dots,w_n(x))^\top := (G_X+n \varepsilon I_n)^{-1} {\bf k}_{\mathcal{X}}(x) \in \mathbb{R}^n, \nonumber
\end{eqnarray}
where $G_X:= (k_{\mathcal{X}}(X_{i},X_{j}))_{i,j=1}^n \in \mathbb{R}^{n \times n}$ is the Gram matrix of $X_1,\dots,X_n$, ${\bf k}_{\mathcal{X}}(x) := (k_{\mathcal{X}}(X_{1},x), \dots k_{\mathcal{X}}(X_{n},x) )^\top \in \mathbb{R}^n$ quantifies the similarities of $x$ and $X_1,\dots,X_n$, $I_n \in \mathbb{R}^{n \times n}$ is the identity matrix, and $\varepsilon > 0$ is a regularization constant.

Noticing that the weight vector $w(x)$ in \eqref{eq:conditionalkernelmean} is identical to that of kernel ridge regression or Gaussian process regression (see e.g.,~\citet[Sect.~3]{KanHenSejSri18}),  one can see that \eqref{eq:conditionalkernelmean} is a regression estimator of the mapping from $x$ to the conditional expectation $\int k_\mathcal{Y}(\cdot,y) p(y|x) dy$. 
This insight has been used by \citet{DBLP:journals/corr/abs-1205-4656} to show that the estimator (\ref{eq:conditionalkernelmean}) is that of {\em function-valued kernel ridge regression}, and to study convergence rates of (\ref{eq:conditionalkernelmean}) by applying results from \citet{CapDev07}.
In the context of structured prediction, \citet{JasonWestonNIPS2003,CorinnaCortesICML2005} derived the same estimator under the name of {\em kernel dependency estimation}, although the connection to embedding of probability distributions was not known at the time. 

\subsection{Nonparametric Kernel Sum Rule (NP-KSR)}
\label{sec:NP-KSR}

Let $\pi(x)$ be a probability density function on $\mathcal{X}$, and $p(y|x)$ be a conditional density function of $y \in \mathcal{Y}$ given $x \in \mathcal{Y}$.
Denote by $q(x,y)$ the joint density defined by $\pi(x)$ and $p(y|x)$:
\begin{equation} \label{eq:joint-density}
q(x,y):=p(y|x)\pi(x), \quad x \in \mathcal{X},\ y \in \mathcal{Y}.
\end{equation}
Then the usual {\em sum rule} is defined as the operation to output the marginal density $q(y)$ on $\mathcal{Y}$ by computing the integral with respect to $x$:
\begin{equation} \label{eq:sum-rule}
q(y) = \int q(x,y) dx = \int p(y|x)\pi(x) dx.     
\end{equation}
For notational consistency, we write the distribution of $q(y)$ as $Q_\mathcal{Y}$. 

The Kernel Sum Rule proposed by \citet{Song+al:icml2010hilbert}, which we call {\em Nonparametric Kernel Sum Rule (NP-KSR)} to distinguish it from the Model-based Kernel Sum Rule proposed in this paper, is an estimator of the kernel mean of the marginal density \eqref{eq:sum-rule}:
\begin{equation} \label{eq:NP-KSR-output}
    m_{Q_{\mathcal{Y}}}:= \int k_\mathcal{Y}(\cdot,y)q(y)dy =  \int \int k_\mathcal{Y}(\cdot,y) p(y|x)\pi(x) dx dy.
\end{equation}
The NP-KSR estimates this using i) training data $\{ (X_i, Y_i)\}_{i=1}^{n} \subset \mathcal{X} \times \mathcal{Y}$ for the conditional density $p(y|x)$ and ii) a weighted sample approximation $\{ (\gamma_i,\tilde{X}_i) \}_{i=1}^\ell \subset \mathbb{R} \times \mathcal{X}$ to the kernel mean $m_{\Pi} := \int k(\cdot,x)\pi(x)dx$ of the input marginal density $\pi(x)$ of the form
\begin{equation} \label{eq:prior-estimate}
\hat m_{\Pi}=\sum_{i=1}^{\ell} \gamma_{i} k_{\mathcal{X}}(\cdot,\tilde{X}_{i}),
\end{equation}
where the subscript $\Pi$ in the left side denotes the distribution of $\pi$. 
To describe the NP-KSR estimator, it is instructive to rewrite \eqref{eq:NP-KSR-output} using the conditional kernel means \eqref{eq:conditionalkernelmeandef} as
$$
m_{Q_{\mathcal{Y}}} =  \int \left( \int k_\mathcal{Y}(\cdot,y) p(y|x) dy \right) \pi(x) dx = \int m_{\mathcal{Y}|x} \pi(x) dx. 
$$
This implies that this kernel mean can be estimated using the estimator (\ref{eq:conditionalkernelmean}) of the conditional kernel means $m_{\mathcal{Y}|x}$ and the weighted sample $\{ (\gamma_i,\tilde{X}_i) \}_{i=1}^\ell$, which can be seen as an empirical approximation of the input distribution 
$\Pi \approx \hat{\Pi} := \sum_{i=1}^\ell \gamma_\ell \delta_{\tilde{X}_i}$,
where $\delta_x$ denotes the Dirac distribution at $x \in \mathcal{X}$.
Thus, the estimator of the NP-KSR is given as
\begin{eqnarray}
\text{\bf NP-KSR}: \hspace{1mm} \hat m_{Q_{\mathcal{Y}}} &:=&  \sum_{i=1}^{\ell} \gamma_{i} \hat m_{\mathcal{Y}|\tilde X_i} = \sum_{j=1}^{n} w_{j} k_{\mathcal{Y}}(\cdot, Y_j),  \nonumber \\
\hspace{1mm} w &:=& (w_1,\dots,w_n)^\top := (G_X+n \varepsilon I_n)^{-1}G_{X\tilde{X}}\gamma,  \label{eq:KernelSumRuleMapping}
\end{eqnarray}  
where $\hat{m}_{\mathcal{Y}|\tilde X_i}$ is \eqref{eq:conditionalkernelmean} with $x = \tilde{X}_i$, $\gamma := (\gamma_1,\dots,\gamma_\ell)^\top \in \mathbb{R}^{\ell}$ and $G_{X\tilde{X}} \in \mathbb{R}^{n \times \ell}$ is such that $(G_{X\tilde{X}})_{i,j} = k_{\mathcal{X}}(X_{i},\tilde{X}_{j})$.
Notice that since $G_{X\tilde{X}}\gamma = ( \sum_{j = 1}^\ell \gamma_i k_\mathcal{X}(X_i,\tilde{X}_j) )_{i=1}^n = (\hat{m}_\Pi(X_i))_{i=1}^n$, the weights in \eqref{eq:KernelSumRuleMapping} can be written as
\begin{equation} \label{eq:weight-NPKSR-eval}
(w_1,\dots,w_n)^\top = (G_X+n \varepsilon I_n)^{-1} (\hat{m}_\Pi(X_1),\dots,\hat{m}_\Pi(X_n))^\top .    
\end{equation}
That is, the weights can be calculated in terms of evaluations of the input empirical kernel mean $\hat{m}_\Pi$ at $X_1,\dots,X_n$; this property will be used in Sect.~\ref{sec:Mb-KSR+NP-KSR}.

The consistency and convergence rates of the estimator \eqref{eq:KernelSumRuleMapping}, which require the regularization constant $\varepsilon$ to decay to $0$ as $n \to \infty$ at an appropriate rate, have been studied in the literature \citep[Theorem 8]{KernelBayes'Rule_BayesianInferencewithPositiveDefiniteKernels}.

\subsection{Kernel Bayes' Rule (KBR)} \label{sec:KBR}
We describe here {\em Kernel Bayes' Rule (KBR)}, an estimator of of the kernel mean of a posterior distribution
\citep{KernelBayes'Rule_BayesianInferencewithPositiveDefiniteKernels}.
Let $\pi(x)$ be a prior density on $\mathcal{X}$ and $p(y|x)$ be a conditional density on $\mathcal{Y}$ given $x \in \mathcal{X}$.
The standard Bayes' rule is an operation to produce the posterior density $q(x|y)$ on $\mathcal{X}$ for a given observation $y \in \mathcal{Y}$ induced from $\pi(x)$ and $p(y|x)$:
$$
q(x|y) = \frac{\pi(x)p(y|x)}{q(y)}, \quad q(y) := \int \pi(x')p(y|x') dx'.  
$$
In the setting of KBR, it is assumed that $\pi(x)$ and $p(y|x)$ are unknown but samples approximating them are available; assume that the prior $\pi(x)$ is approximated by weighted points $\{(\gamma_i,\tilde{X}_i)\}_{i=1}^\ell \subset \mathbb{R} \times \mathcal{X}$ in the sense that its kernel mean $m_\Pi := \int k_\mathcal{X}(\cdot,x)\pi(x)dx$ is approximated by $\hat m_{\Pi} := \sum_{i=1}^{\ell} \gamma_{i} k_{\mathcal{X}}(\cdot,\tilde{X}_{i})$ as in \eqref{eq:prior-estimate}, and that training data $\{(X_i,Y_i)\}_{i=1}^n \subset \mathcal{X} \times \mathcal{Y}$ are provided for the conditional density $p(y|x)$. 
Using $\hat m_{\Pi}$ and $\{(X_i,Y_i)\}_{i=1}^n$, the KBR estimates the kernel mean of the posterior 
$$
{{m}}_{Q_{\mathcal{X}| y} } := \int k_\mathcal{X}(\cdot,x) q(x|y) dx.
$$


Specifically the estimator of the KBR is given as follows.
Let $w \in \mathbb{R}^n$ be the weight vector defined as (\ref{eq:KernelSumRuleMapping}) or \eqref{eq:weight-NPKSR-eval}, and $D(w) \in \mathbb{R}^{n \times n}$ be a diagonal matrix with its diagonal elements being $w$.
Then the estimator of the KBR is defined by
\begin{eqnarray}
\mathbf{KBR}: \hspace{1mm} {\hat{m}}_{Q_{\mathcal{X}| y} } &=&
\sum_{j=1}^{n} \tilde{w} _{j} k_{\mathcal{X}}(\cdot, X_j), \hspace{3mm}  \tilde{w} := R_{ \mathcal{X| Y}} {\bf k}_{\mathcal{Y}}(y) \in \mathbb{R}^n,  \nonumber \\
 {R_{\mathcal{X| Y}}} &\!:=\!& D(w){G_Y}{ \left( {{{\left( {D(w){G_Y}} \right)}^2} + \delta {I_n}} \right )^{ - 1}} \hspace{-1mm}D(w) \in \mathbb{R}^{n \times n},
 \label{eq:kernelBayesrule}
\end{eqnarray}
where ${\bf k}_{\mathcal{Y}}(y):=(k_{\mathcal{Y}}(y, Y_1), \ldots, k_{\mathcal{Y}}(y, Y_n))^{\top} \in \mathbb{R}^{n}$, $G_Y = (k_{\mathcal{Y}}(Y_i,Y_j) )_{i,j=1}^n \in \mathbb{R}^{n \times n}$, and $\delta > 0$ is a regularization constant. 
This is a consistent estimator: As the number of training data $n$ increases and as $\hat{m}_\Pi$ approaches $m_\Pi$, the estimate $ {\hat{m}}_{Q_{\mathcal{X}| y} }$ converges to ${{m}}_{Q_{\mathcal{X}| y} }$ under certain assumptions; see \citet[Theorems 6 and 7]{KernelBayes'Rule_BayesianInferencewithPositiveDefiniteKernels} for details.

\section{Kernel Bayesian Inference with Probabilistic Models} \label{sec:KernelBayesianInferencewithProbabilisticModels}
In this section, we introduce the Model-based Kernel Sum Rule (Mb-KSR), a realization of the sum rule in kernel Bayesian inference using a probabilistic model. 
We describe the Mb-KSR in Sect. \ref{sec:ModelbasedKernelSumRule(Mb-KSR)}, and show how to combine the MB-KSR and NP-KSR in Sect. \ref{sec:CombinationofNP-KSRandMb-KSR}.
We explain how the KBR can be implemented when a prior kernel mean estimate is given by a model-based algorithm such as the Mb-KSR in Sect. \ref{sec:KBRpriorprobabilisticmodel}.
We will use these basic estimators to develop a filtering algorithm for state space models in Sect. \ref{sec:FilteringinStateSpaceModels}.  
As mentioned in Sect.~\ref{sec:conditional-mean-embeddings}, we assume that distributions under considerations have density functions for the sake of clarity of presentation.

\subsection{Model-based Kernel Sum Rule (Mb-KSR)} \label{sec:ModelbasedKernelSumRule(Mb-KSR)} 
Let $\mathcal{X} = \mathcal{Y} = \mathbb{R}^m$ with $m \in \mathbb{N}$. 
Define kernels $k_\mathcal{X}$ and $k_\mathcal{Y}$ on $\mathcal{X}$ and $\mathcal{Y}$, respectively, and let $\mathcal{H}_\mathcal{X}$ and $\mathcal{H}_\mathcal{Y}$ be their respective RKHSs.
Assume that a user defines a probabilistic model as a conditional density function\footnote{For simplicity of presentation we assume the probabilistic model has a density function, but the framework below can also hold even when this assumption does not hold (e.g., when the mapping $x \to y$ is deterministic, in which case the conditional distribution is given with a Dirac delta function).} on $\mathcal{Y}$ given $\mathcal{X}$:
$$
p_M(y|x), \quad x,y \in \mathbb{R}^m,
$$ 
where the subscript ``$M$'' stands for ``Model.''
Consider the kernel mean of the probabilistic model $p_M(y|x)$:
\begin{equation} \label{eq:cond-mean-model}
    m_{\mathcal{Y}| x} = \int k_\mathcal{Y}(\cdot,y) p_M(y|x) dy \in \mathcal{H_Y}, \quad x \in \mathcal{X}.
\end{equation}
We focus on situations where the above integral has an analytic solution, and thus one can evaluate the value of the kernel mean $m_{\mathcal{Y}|x}(y') =  \int k_\mathcal{Y}(y',y) p_M(y|x) dy$ for a given $y' \in \mathcal{Y}$.

An example is given by the case where $p_M(y|x)$ is an additive Gaussian noise model, as described in Example \ref{ex:conditionalGaussiankernelmean} below. (Other examples can be found in Appendix \ref{sec:Appendix}.)
To describe this, let $N(\mu, R)$ be the $m$-dimensional Gaussian distribution with mean vector $\mu \in \mathbb{R}^m$ and covariance matrix $R \in \mathbb{R}^{m \times m}$, and let $g(x|\mu,R)$ denote its density function:
\begin{equation} \label{eq:Gauss-density}
    g(x|\mu,R) := |2\pi R|^{-1/2} \ \exp\left( (x-\mu)^\top R^{-1} (x-\mu) \right).
\end{equation}
Then an additive Gaussian noise model is such that an output random variable $Y \in \mathbb{R}^m$ conditioned on an input $x \in \mathcal{X}$ is given as
$$
Y=f(x)+\epsilon, \quad \epsilon \sim N(0,\Sigma),
$$ 
where $f\colon\mathcal{X} \rightarrow \mathbb{R}^{m}$ is a vector-valued function and $\Sigma \in \mathbb{R}^{m \times m}$ is a covariance matrix; or equivalently, the conditional density function is given as
\begin{equation} \label{eq:additive-Gauss}
p_M(y | x) = g(y| f(x), \Sigma), \quad x,y \in \mathbb{R}^m.
\end{equation}
The additive Gaussian noise model is ubiquitous in the literature, since the form of the Gaussian density often leads to convenient analytic expressions for quantities of interest.
An illustrative example is the Kalman filter \citep{Kal60}, which uses linear-Gaussian models for filtering in state space models; in the notation of \eqref{eq:additive-Gauss}, this corresponds to $f$ being a linear map. 
Another example is Gaussian process models \citep{GaussianProcessesforMachineLearning}, for which additive Gaussian noises are often assumed with $f$ being a nonlinear function following a Gaussian process.


The following describes how the conditional kernel means can be calculated for additive Gaussian noise models by using Gaussian kernels.
\begin{Example}[An additive Gaussian noise model with a Gaussian kernel] \label{ex:conditionalGaussiankernelmean}
Let $p_M(y|x)$ be an additive Gaussian noise model defined as \eqref{eq:additive-Gauss}.
For a positive definite matrix $R \in \mathbb{R}^{m \times m}$, let $k_R: \mathbb{R}^m \times \mathbb{R}^m \to \mathbb{R}$ be a normalized Gaussian kernel \footnote{Here we use a normalized Gaussian kernel $k_{R}(x_{1},x_{2})=g(x_{1}-x_{2}| 0,R)$ that is of the form of a probability density function, rather than the unnormalized kernel $\tilde k_{R}(x_1, x_2) = \exp (-\frac{1}{2}(x_1-x_2)^{\top}R^{-1}(x_1-x_2))$ standard in the literature \citep[p. 153]{SupportVectorMachines2008}. 
Our motivation is that, if the normalized kernel is used, then the kernel mean is also of the form of a probability density function, which is convenient since the coefficient is not required to be adjusted. 
On the other hand, if the unnormalized kernel $\tilde k_{R}$ is used, then the resulting kernel mean should be multiplied by a constant as $\tilde m _{\mathcal{Y}| x} =  | 2 \pi R|^{1/2} m _{\mathcal{Y}| x}$, where $| 2 \pi R|^{1/2}$ is the normalization constant of the Gaussian probability density. 
We use normalized kernels also for other noise models; see Appendix \ref{sec:Appendix}.} defined as
\begin{equation} \label{eq:gaussian-kernel}
    k_{R}(x_{1},x_{2})=g(x_{1}-x_{2}| 0,R), \quad x_1,x_2 \in \mathbb{R}^m,
\end{equation}
where $g$ is the Gaussian density \eqref{eq:Gauss-density}.
  Then the conditional kernel mean  \eqref{eq:cond-mean-model} with $k_\mathcal{Y} := k_{R}$ is given by
\begin{eqnarray}
{m _{\mathcal{Y}| x}} (y) = g( y
| f(x),\Sigma + R ), \quad x,y \in \mathbb{R}^m. \label{eq:conditionalGaussianAnalyticalkernelmeannaive}
\end{eqnarray}
\end{Example}

\begin{proof}
For each $x \in \mathcal{X}$, the conditional kernel mean \eqref{eq:cond-mean-model} can be written in the form of convolution, ${m _{\mathcal{Y}| x}}(y)  = \int  g(y - y' | 0 ,R) g( y' | f(x),\Sigma) dy' =: g(\cdot | 0, R) * g(\cdot| f(x), \Sigma) (y)$. 
and \eqref{eq:conditionalGaussianAnalyticalkernelmeannaive}  follows from the well-known fact that the convolution of two Gaussian probability densities is given by $g(\cdot | \mu_1,\Sigma_1)*g(\cdot | \mu_2,\Sigma_2) = g(\cdot | \mu_1+\mu_2,\Sigma_1+\Sigma_2)$.
\end{proof}


As in Sect.~\ref{sec:NP-KSR}, let $\pi(x)$ be a probability density function on $\mathcal{X}$ and define the marginal density $q(y)$ on $\mathcal{Y}$ by
$$
q(y) = \int p_M(y|x) \pi(x) dx, \quad y \in \mathcal{Y}.
$$
The Mb-KSR estimates the kernel mean of this marginal probability 
\begin{equation} \label{eq:Mb-KSR-objective}
m_{Q_\mathcal{Y}} :=  \int k_\mathcal{Y}(\cdot,y)q(y)dy = \int\left( \int  k_\mathcal{Y} (\cdot,y) p_M(y|x) dy \right) \pi(x) dx. 
\end{equation}
This is done by using the probabilistic model $p_M(y|x)$ and an empirical approximation $\hat m_{\Pi}=\sum_{i=1}^{\ell} \gamma_{i} k_{\mathcal{X}}(\cdot,\tilde{X}_{i})$ to the kernel mean $m_\Pi = \int k_\mathcal{X} (\cdot,x)\pi(x)dx$ of the input probability $\pi(x)$. 
Since the weighted points $\{(\gamma_i,\tilde{X}_i)\}_{i=1}^\ell \subset \mathbb{R} \times \mathcal{X}$ provide an approximation to the distribution $\Pi$ of $\pi$ as $\Pi \approx \hat{\Pi} := \sum_{i=1}^\ell \gamma_\ell \delta_{\tilde{X}_i}$, we define the Mb-KSR as follows:
\begin{equation}
\text{\bf Mb-KSR}:  \hspace{10mm} {\hat m _{Q_{\mathcal{Y}} }} := \sum_{i=1}^{\ell} \gamma_{i} m _{\mathcal{Y}|  \tilde{X}_i } = \sum_{i=1}^\ell \gamma_i \int k_\mathcal{Y}(\cdot,y) p_M(y|\tilde{X}_i) dy ,
\label{eq:MbKSRestimator}
\end{equation}
where $m _{\mathcal{Y}|  \tilde{X}_i }$ is the conditional kernel mean \eqref{eq:cond-mean-model} with $x := \tilde{X}_i$.
In the case of Example \ref{ex:conditionalGaussiankernelmean}, for instance, 
one can compute the value ${\hat m _{Q_{\mathcal{Y}} }}(y)$ for any given $y \in \mathcal{Y}$ by using the analytic expression (\ref{eq:conditionalGaussianAnalyticalkernelmeannaive}) of $m _{\mathcal{Y}|  \tilde{X}_i }$ in (\ref{eq:MbKSRestimator}).
As mentioned earlier, however, one can use for the Mb-KSR other noise models by employing appropriate kernels, as described in Appendix \ref{sec:Appendix}.
One such example is an additive Cauchy noise model with a rational quadratic kernel \citep[Eq.~4.19]{GaussianProcessesforMachineLearning}, which should be useful when modeling heavy-tailed random quantities.

We provide convergence results of the Mb-KSR estimator (\ref{eq:MbKSRestimator}), as shown in Proposition \ref{Prop:consistencyMb-KSR} below.
The proof can be found in Appendix \ref{subsec:consistency}. 
Below $O_p$ is the order notation for convergence in probability, and $\mathcal{H_X} \otimes \mathcal{H_X}$ denotes the tensor product of two RKHSs $\mathcal{H_X}$ and $\mathcal{H_X}$.

\begin{Proposition} \label{Prop:consistencyMb-KSR}
Let $\{(\gamma_i,\tilde{X}_i)\}_{i=1}^\ell \subset \mathbb{R} \times \mathcal{X}$ be such that $\hat m_{\Pi} := \sum_{i=1}^{\ell}\gamma_i k_{\mathcal{X}}(\cdot, \tilde{X}_i)$, satisfies $|| \hat m_{\Pi}-m_{\Pi} ||_{\mathcal{H_X}}=O_p(\ell^{-\alpha})$ as $\ell \rightarrow \infty$ for some $\alpha > 0$. 
For a function $\theta: \mathcal{X} \times \mathcal{X} \to \mathbb{R}$ defined by
$\theta (x,\tilde x): = \int \int {k_{\mathcal{Y}}}(y,\tilde y) p_M(y|x) p_M(\tilde{y}|\tilde{x}) dy d\tilde{y}$ for $(x, \tilde{x}) \in \mathcal{X} \times \mathcal{X}$, assume that $\theta \in \mathcal{H_X} \otimes \mathcal{H_X}$. 
Then for $m_{{Q_\mathcal{Y}}}$ and $\hat m_{Q_\mathcal{Y}}$ defined respectively in \eqref{eq:Mb-KSR-objective} and \eqref{eq:MbKSRestimator}, we have
$$
\left\| {{m_{{Q_\mathcal{Y}}} } - \hat m_{Q_\mathcal{Y}} } \right\|_{{\mathcal{H}_\mathcal{Y}}} = O_p({\ell^{ - \alpha }})  \quad ( \ell \to \infty).
$$
\end{Proposition}

\begin{remark}
The convergence rate of $\hat m_{{Q_{\mathcal{Y}}}}$ given by the Mb-KSR in Proposition \ref{Prop:consistencyMb-KSR} is the same as that of the input kernel mean estimator $\hat m_{\Pi}$.
On the other hand, the rate for the NP-KSR is known to become slower than that of the input estimator, because of the need for additional learning and regularization \citep[Theorem 8]{KernelBayes'Rule_BayesianInferencewithPositiveDefiniteKernels}. 
Therefore Proposition \ref{Prop:consistencyMb-KSR} shows an advantage of the Mb-KSR over the NP-KSR, when the probabilistic model is correctly specified.
The condition that $\theta(\cdot, \cdot) \in \mathcal{H_X} \otimes \mathcal{H_X}$ is the same as the one made in \citet[Theorem 8]{KernelBayes'Rule_BayesianInferencewithPositiveDefiniteKernels}.  
\end{remark}


For any function of the form $f = \sum_{j=1}^{m} c_j k_{\mathcal{Y}}(\cdot, y_j) \in \mathcal{H_Y}$ with  $c_1,\dots,c_m \in \mathbb{R}$ and $y_1,\dots,y_m \in \mathcal{Y}$, its expectation with respect to $q(y)$ can be approximated using the Mb-KSR estimator \eqref{eq:MbKSRestimator} as
\begin{eqnarray}
\int f(y) q(y)dy  = \sum_{j=1}^m c_j m_{Q_\mathcal{Y}}(y_j) \approx  \sum_{j=1}^{m} c_j \sum_{i=1}^{\ell} \gamma_{i} m_{\mathcal{Y}|\tilde X_i}(y_j)  . \label{eq:expectationestimateMbKSR}
\end{eqnarray}



\subsection{Combining the Mb-KSR and NP-KSR}
\label{sec:CombinationofNP-KSRandMb-KSR}

\begin{figure}[t]
\begin{center}
\includegraphics[width=0.9\linewidth]{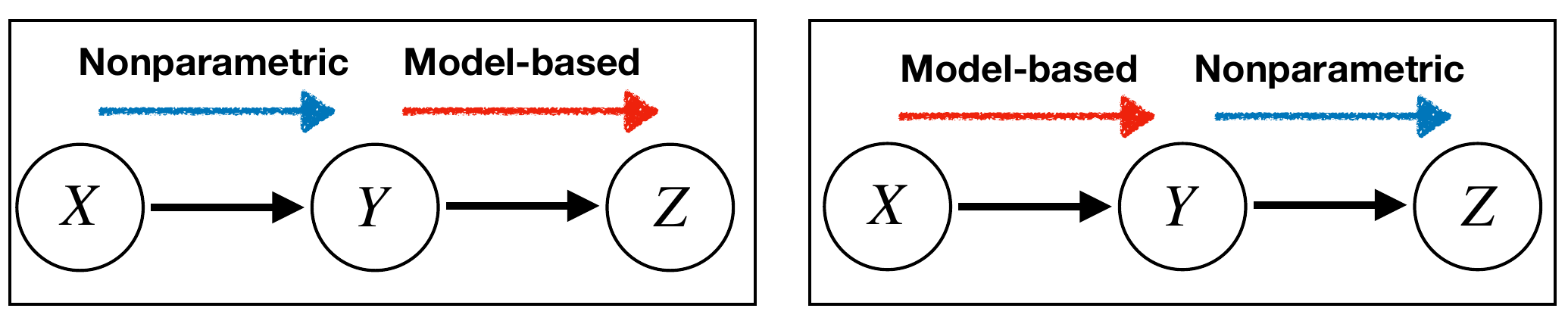}
\caption{Hybrid kernel Bayesian inference in a three-variables chain graphical model.}
\label{fig:three-variable1}
\end{center}
\end{figure}

Using the Mb-KSR and NP-KSR, one can perform hybrid (i.e., model-based and nonparametric) kernel Bayesian inference. 
In the following we describe two examples of such hybrid inference with a simple chain graphical model (Fig. \ref{fig:three-variable1}). 
In Sect.~\ref{sec:FilteringinStateSpaceModels}, we use the estimators derived below corresponding to the two figures in Fig. \ref{fig:three-variable1} to develop our filtering algorithm for state space models.

To this end, let $\mathcal{X}$, $\mathcal{Y}$, and $\mathcal{Z}$ be three measurable spaces, and let $k_\mathcal{X}$, $k_\mathcal{Y}$ and $k_\mathcal{Z}$ be kernels defined on the respective spaces.
For both of the two cases below, let $\pi(x)$ be a probability density function on $\mathcal{X}$.
Assume that we are given weighted points $\{ (w_i,\tilde{X}_i) \}_{i=1}^\ell \subset \mathbb{R} \times \mathcal{X}$ that provide an approximation $\hat m_{\Pi}=\sum_{i=1}^{\ell} \gamma_i k_{\mathcal{X}}(\cdot, \tilde{X}_i)$ to the kernel mean $\int k_\mathcal{X}(\cdot,x)\pi(x)dx$.


\subsubsection{NP-KSR followed by Mb-KSR (Fig. \ref{fig:three-variable1}, Left)}
\label{sec:NP-KSR+Mb-KSR}

Let $p(y|x)$ be a conditional density function of $y \in \mathcal{Y}$ given $x \in \mathcal{X}$, and $p_M(z|y)$ be a conditional  density function of $z \in \mathcal{Z}$ given $y \in \mathcal{Y}$.
Suppose that $p(y|x)$ is unknown, but training data $\{(X_{i},Y_i)\}_{i=1}^{n} \subset \mathcal{X} \times \mathcal{Y}$ for it are available. 
On the other hand, $p_M(z|y)$ is a probabilistic model, and assume that the kernel $k_\mathcal{Z}$ is chosen so that the conditional kernel mean $m_{\mathcal{Z}|y} := \int k_\mathcal{Z}(\cdot,z)p_M(z|y)$ is analytically computable for each $y \in \mathcal{Y}$.
Define marginal densities $q(y)$ on $\mathcal{Y}$ and $q(z)$ on $\mathcal{Z}$ by
$$
q(y) := \int \pi(x) p(y|x) dx , \quad q(z) :=  \int q(y) p_M(z|y) dy,
$$ 
and let $m_{Q_{\mathcal{Y}}} := \int k_\mathcal{Y}(\cdot,y)q(y) dy$ and  $m_{Q_{\mathcal{Z}}} := \int k_\mathcal{Z}(\cdot,z)q(z)dz$ be their respective kernel means.

The goal here is to estimate  $m_{Q_{\mathcal{Z}}}$ using  $\hat m_{\Pi}=\sum_{i=1}^{\ell} \gamma_i k_{\mathcal{X}}(\cdot, \tilde{X}_i)$,  $\{(X_{i},Y_i)\}_{i=1}^{n}$ and $p_M(z|y)$.
This can be done by two steps: 
i) first estimate the kernel mean $m_{Q_{\mathcal{Y}}}$ using the NP-KSR \eqref{eq:KernelSumRuleMapping} with $\hat m_{\Pi}$ and $\{(X_{i},Y_i)\}_{i=1}^{n}$, obtaining an estimate $\hat m_{Q_{\mathcal{Y}}} =  \sum_{j=1}^{n} w_{j} k_{\mathcal{Y}}(\cdot, Y_j)$ with $ w := (w_1,\dots,w_n)^\top := (G_X+n \varepsilon I_n)^{-1}G_{X\tilde{X}}\gamma$, where  $\gamma := (\gamma_1,\dots,\gamma_\ell)^\top \in \mathbb{R}^{\ell}$, $G_{X\tilde{X}} \in \mathbb{R}^{n \times \ell}$ is such that $(G_{X\tilde{X}})_{i,j} = k_{\mathcal{X}}(X_{i},\tilde{X}_{j})$ and $\varepsilon > 0$ is a regularization constant;
then ii) apply the Mb-KSR to $\hat m_{Q_{\mathcal{Y}}}$ using $p_M(z|y)$, resulting in the following estimator of $m_{Q_{\mathcal{Z}}}$:
\begin{eqnarray}
\hat m_{Q_{\mathcal{Z}}} = \sum_{i=1}^{n} w_{i} m_{\mathcal{Z}| Y_i}, \quad \text{where }\   m_{\mathcal{Z}|Y_i} := \int k_\mathcal{Z}(\cdot,z)p_M(z|Y_i) dz.
\label{eq:typeikernelmeanestimation}
\end{eqnarray}

\subsubsection{Mb-KSR followed by NP-KSR (Fig. \ref{fig:three-variable1}, Right)} 
\label{sec:Mb-KSR+NP-KSR}
 
Let $p_M(y|x)$ be a conditional density function of $y \in \mathcal{Y}$ given $x \in \mathcal{X}$, and $p(z|y)$ be a conditional density function of $z \in \mathcal{Z}$ given $y \in \mathcal{Y}$.
Suppose that for the probabilistic model $p_M(y|x)$, the kernel $k_\mathcal{Y}$ is chosen so that the conditional kernel mean $m_{\mathcal{Y}|x} := \int k_\mathcal{Y}(y|x)dy$ is analytically computable for each $x \in \mathcal{X}$.
On the other hand, assume that training data $\{(Y_{i},Z_i)\}_{i=1}^{n} \subset \mathcal{Y} \times \mathcal{Z}$ for the unknown conditional density $p(z|y)$ are available. 
Define marginal densities $q(y)$ on $\mathcal{Y}$ and $q(z)$ on $\mathcal{Z}$ by
$$
q(y) := \int \pi(x) p_M(y|x) dx , \quad q(z) :=  \int q(y) p(z|y) dy,
$$ 
and let $m_{Q_{\mathcal{Y}}} := \int k_\mathcal{Y}(\cdot,y)q(y) dy$ and  $m_{Q_{\mathcal{Z}}} := \int k_\mathcal{Z}(\cdot,z)q(z)dz$ be their respective kernel means.

The task is to estimate  $m_{Q_{\mathcal{Z}}}$ using  $\hat m_{\Pi}=\sum_{i=1}^{\ell} \gamma_i k_{\mathcal{X}}(\cdot, \tilde{X}_i)$, $p_M(y|x)$ and $\{(Y_{i},Z_i)\}_{i=1}^{n} \subset \mathcal{Y} \times \mathcal{Z}$. 
This can be done by two steps:
i) first estimate the kernel mean $m_{Q_{\mathcal{Y}}}$ by applying the Mb-KSR \eqref{eq:MbKSRestimator} to $\hat m_{\Pi}$, yielding an estimate ${\hat m _{Q_{\mathcal{Y}} }} := \sum_{i=1}^{\ell} \gamma_{i} m _{\mathcal{Y}|  \tilde{X}_i }$, where $m _{\mathcal{Y}|  \tilde{X}_i } = \int k_\mathcal{Y}(\cdot,y) p_M(y|\tilde{X}_i) dy$;
ii) then apply the NP-KSR to ${\hat m _{Q_{\mathcal{Y}} }}$.
To describe ii), recall that the weights for the NP-KSR can be written as \eqref{eq:weight-NPKSR-eval} in terms of evaluations of the input empirical kernel mean: thus, the estimator of $m_{Q_{\mathcal{Z}}}$ by the NP-KSR in ii) is given by
\begin{eqnarray}
\hat m_{Q_{\mathcal{Z}}} = 
\sum_{i=1}^{n} w_i k_{\mathcal{Z}}(\cdot, Z_i), \label{eq:typeiicase}
\end{eqnarray}
with the weights $w_1,\dots,w_n$ being
\begin{eqnarray*}
(w_1,\dots,w_n)^\top &:=& (G_Y+n \varepsilon I_n)^{-1} ({\hat m _{Q_{\mathcal{Y}} }}(Y_1), \dots, {\hat m _{Q_{\mathcal{Y}} }} (Y_n)) )^\top \\
&=& (G_Y+n \varepsilon I_n)^{-1}G_{Y| \tilde{X}}\gamma,
\end{eqnarray*}
where $G_{Y| \tilde{X}} \in \mathbb{R}^{n \times \ell}$ is such that $(G_{Y| \tilde{X}})_{ij}={{m_{\mathcal{Y}| {{\tilde{X}}_j} }}({{Y}_i})} = \int k_\mathcal{Y}(Y_i,y) p_M(y|\tilde{X}_i) dy$ and $\gamma := (\gamma_1,\dots,\gamma_\ell)^\top \in \mathbb{R}^\ell$. 

\subsection{Kernel Bayes' Rule with a Model-based Prior} \label{sec:KBRpriorprobabilisticmodel} 
We describe how the KBR in Sect.~\ref{sec:KBR} can be used when the prior kernel mean $\hat{m}_{\Pi}$ is given by a model-based estimator such as (\ref{eq:MbKSRestimator}).
This way of applying KBR is employed in Sect. \ref{sec:FilteringinStateSpaceModels} to develop our filtering method. 
The notation in this subsection follows that in Sect.~\ref{sec:KBR}.

Denote by $\hat m _{\Pi} := \sum_{j=1}^{\ell} \gamma_{j} m _{j}$ a prior kernel mean estimate, where  $m_1,\dots,m_\ell \in \mathcal{H_X}$ represent model-based kernel mean estimates and $\gamma_1,\dots,\gamma_\ell \in \mathbb{R}$; for later use, we have written the kernel means $m_1,\dots,m_\ell$ rather abstractly.
For instance, if $\hat m _{\Pi}$ is obtained from the Mb-KSR \eqref{eq:MbKSRestimator}, then
$m_{j}$ may be given in the form $m_{j} = \int k_\mathcal{X}(\cdot,x) p_M(x|\tilde{X}_j) dx$ for some probabilistic model $p_M(x|\tilde{x})$ and some $\tilde{X}_j \in \mathcal{X}$.

Then the KBR with the prior $\hat m _{\Pi}$ is simply given by the estimator (\ref{eq:kernelBayesrule}) with the weight vector $w \in \mathbb{R}^n$ replaced by the following:
$$
w = {\left( {{G_X} + n\varepsilon {I_n}} \right)^{ - 1}}{M \gamma} \in \mathbb{R}^n, 
$$
where $\gamma := (\gamma_1,\dots,\gamma_n)^\top \in \mathbb{R}^\ell$ and $M \in \mathbb{R}^{n \times \ell}$ is such that $M_{ij} = m_{j}(X_i)$. 
This follows from that the weight vector $w$ for the KBR is that of the NP-KSR \eqref{eq:weight-NPKSR-eval}; see also Sect.~\ref{sec:Mb-KSR+NP-KSR}.

\section{Filtering in State Space Models via Hybrid Kernel Bayesian Inference} \label{sec:FilteringinStateSpaceModels}
Based on the framework for hybrid kernel Bayesian inference introduced in Sect. \ref{sec:KernelBayesianInferencewithProbabilisticModels}, we propose a novel filtering algorithm for state space models, focusing on the setting of Fig.~\ref{fig:state-space-model}.
We formally state the problem setting in Sect.~\ref{sec:settingSSM}, and then describe the proposed algorithm in Sect.~\ref{sec:Mc-KBFilterAlgorithm}, followed by an explanation about how to use the outputs of the proposed algorithm in Sect.~\ref{sec:ComputationfromOutput}.
As before, we assume that all distributions under consideration have density functions for clarity of presentation.

\subsection{The Problem Setting} \label{sec:settingSSM}
Let $\mathcal{X}$ be a space of states, and $\mathcal{Z}$ be a space of observations.
Let $t=1,\dots,T$ denotes the time index with $T \in \mathbb{N}$ being the total number of time steps.
A state space model (Fig.~\ref{fig:state-space-model}) consists of two kinds of variables: states $x_1, x_2, \dots,x_T \in \mathcal{X}$ and observations $z_1,z_2,\dots,z_T \in \mathcal{Z}$.
These variables are assumed to satisfy the conditional independence structure described in Fig.~\ref{fig:state-space-model}, and probabilistic relationships between the variables are specified by two conditional density functions: 1) a transition model $p(x_{t+1} | x_t)$ that describes how the next state $x_{t+1}$ can change from the current state $x_t$; and 2) an observation model $p(z_t | x_t)$ that describes how likely the observation $z_t$ is generated from the current state $x_t$.
Let $p(x_1)$ be a prior of the initial state $x_1$.

In this paper, we focus on the case where the transition process is an additive Gaussian noise model, which has been frequently used in the literature.
As mentioned before, nevertheless, other noise models described in Appendix \ref{sec:ConditionalKernelMeanofProbabilisticModel} can also be used. 
We consider the following setting.
\begin{itemize}
\item {\bf Transition Model}: 
Let $\mathcal{X} = \mathbb{R}^{m}$, and $k_\mathcal{X}$ be a Gaussian kernel of the form \eqref{eq:gaussian-kernel} with covariance matrix $R \in \mathbb{R}^{n \times n}$.
Define a vector-valued function $f: \mathbb{R}^m \to \mathbb{R}^m$ and a covariance matrix $\Sigma \in \mathbb{R}^{n \times n}$.
It is assumed that $f$ and $\Sigma$ are provided by a user, and thus known.
The transition model is an additive Gaussian noise model such that $x_{t+1}=f(x_{t})+{\epsilon _t}$ with $\epsilon_t \sim N({\bf 0}, \Sigma)$, or in the density from,
$$p(x_{t+1}|x_t) = g(x_{t+1} | f(x_t), \Sigma),$$
where $g(x|\mu,R)$ denotes the Gaussian density with mean $\mu \in \mathbb{R}^m$ and covariance matrix $R \in \mathbb{R}^{m \times m}$; see \eqref{eq:Gauss-density}.

\item {\bf Observation Model}: 
Let $\mathcal{Z}$ be an arbitrary domain on which a kernel $k_{\mathcal{Z}}: \mathcal{Z} \times \mathcal{Z} \to \mathbb{R}$ is defined. 
We assume that training data 
$$\{ (X_i,Z_i) \}_{i=1}^{n} \subset \mathcal{X} \times \mathcal{Z}$$ 
are available for the observation model $p(z_t|x_t)$.
The user is not required to have knowledge about the form of $p(z_t|x_t)$.
\end{itemize}

The task of filtering is to compute the posterior $p(x_t|z_{1:t})$ of the current state $x_t$ given the history of observations $z_{1:t} := (z_1,\dots,z_t)$ obtained so far; this is to be done sequentially for all time steps $t = 1,\dots, T$.
In our setting, one is required to perform filtering on the basis of the transition model $p(x_{t+1}|x_t)$ and the training data $\{ (X_i,Z_i) \}_{i=1}^{n}$.

Regarding the setting above, note that the training data $\{ (X_i,Z_i)\}_{i=1}^{n}$ are assumed to be available {\em before} the test phase.
This setting appears when {\em directly measuring the states of the system is possible but requires costs (in terms of computations, time or money) much higher than those for obtaining observations.}
For example, in the robot localization problem discussed in Sect.~\ref{sec:intro-combination}, it is possible to measure the positions of a robot by using an expensive radar system or by manual annotation, but in the test phase the robot may only be able to use cheap sensors to obtain observations, such as camera images and signal strength information \citep{pronobis2009ijrr}.
Another example is problems where states can be accurately estimated or recovered from data only available before the test phase. 
For instance, in tsunami studies (see e.g.~\citealt{Sai19}), one can recover a tsunami in the past on the basis of data obtained from various sources; however in the test phase, where the task may be that of early warming of a tsunami given that an earthquake has just occurred in an ocean, one can only make use of observations from limited sources, such as seismic intensities and ocean-bottom pressure records.


\subsection{The Proposed Algorithm} \label{sec:Mc-KBFilterAlgorithm}
In general, a filtering algorithm for a state space model consists of two steps: the {\it prediction step} and the {\it filtering step}.
We first describe these two steps, as this will be useful in understanding the proposed algorithm.

Assume that the  posterior $p(x_{t-1} | z_{1:t-1})$ at time $t-1$ has already been obtained. 
(If $t = 1$, start from the filtering step below, with $p(x_1|z_{1:0}) := p(x_1)$)
In the prediction step, one computes the predictive density $p(x_{t} | z_{1:t-1})$ by using the sum rule with the transition model $p(x_t|x_{t-1})$: 
$$
p(x_{t} | z_{1:t-1}) = \int p(x_t | x_{t-1})p(x_{t-1} | z_{1:t-1})dx_{t-1}.
$$
Suppose then that a new observation $z_t$ has been provided.
In the filtering step, one computes the  posterior $p(x_{t} | z_{1:t})$ by using Bayes' rule with $p(x_{t} | z_{1:t-1})$ as a prior and the observation model $p(z_t|x_t)$ as a likelihood function:
$$
p(x_{t} | z_{1:t}) \propto p(z_t| x_t) p(x_{t} | z_{1:t-1}) 
$$
Iterations of these two steps over times $t=1,\dots,T$ result in a filtering algorithm.

We now describe the proposed algorithm. 
In our approach, the task of filtering is formulated as estimation of the kernel mean of the posterior $p(x_t|z_{1:t})$:
\begin{equation} \label{eq:filtering-posterior-mean}
     {m _{{\mathcal{X}_{t}}| {{z}_{1:t}} }} := \int k_{\mathcal{X}}(\cdot, x_{t}) p(x_{t} | z_{1:t}) dx_{t} \in \mathcal{H}_{\mathcal{X}},
\end{equation}
which is to be done sequentially for each time  $t = 1,\dots,T$ as a new observation $z_t$ is obtained. (Here $\mathcal{H}_{\mathcal{X}}$ is the RKHS of $k_\mathcal{X}$.)
The prediction and filtering steps of the proposed algorithm are defined as follows.

\begin{description}
\item[{\bf Prediction Step:}] 
Let ${m _{{\mathcal{X}_{t-1}}| {{z}_{1:t-1}} }} \in \mathcal{H}_\mathcal{X}$ be the kernel mean of the posterior $p(x_{t-1}| z_{1:t-1})$ at time $t-1$
$$
{m _{{\mathcal{X}_{t-1}}| {{z}_{1:t-1}} }} := \int k_{\mathcal{X}}(\cdot, x_{t-1}) p(x_{t-1} | z_{1:t-1}) dx_{t-1} , 
$$
and assume that its estimate ${\hat{m} _{{\mathcal{X}_{t-1}}| {z_{1:t-1}} }} \in \mathcal{H}_\mathcal{X}$ has been computed in the form
\begin{equation} \label{eq:prior-previous}
{\hat{m} _{{\mathcal{X}_{t-1}}| {z_{1:t-1}} }} = \sum_{i=1}^{n} [\boldsymbol{\alpha}_{{\mathcal{X}_{t-1}}| {{z}_{1:t-1}} }]_i k_{\mathcal{X}}(\cdot, X_i), \quad \text{where }\ \boldsymbol{\alpha}_{{\mathcal{X}_{t-1}}| {{z}_{1:t-1}} }  \in \mathbb{R}^{n},    
\end{equation}
where $X_1, \dots,X_n$ are those of the training data.
(If $t=1$, start from the filtering step below.)
The task here is to estimate the kernel mean of the predictive density $p(x_{t} | z_{1:t-1})$:
$$
{m _{{\mathcal{X}_{t}}| {{z}_{1:t-1}} }} := \int k_{\mathcal{X}}(\cdot, x_{t}) p(x_{t} | z_{1:t-1}) dx_{t}.
$$
To this end, we apply the Mb-KSR  (Sect.~\ref{sec:ModelbasedKernelSumRule(Mb-KSR)}) to \eqref{eq:prior-previous} using the transition model $p(x_t|x_{t-1})$ as a probabilistic model: the estimate is given as
\begin{eqnarray}
{\hat m _{{\mathcal{X}_{t}}| {{z}_{1:t-1}} }} &:=&  \sum_{i=1}^{n} [\boldsymbol{\alpha}_{{\mathcal{X}_{t-1}}| {{z}_{1:t-1}} }]_i m_{\mathcal{X}_t| x_t = X_i},  \label{eq:forwardkernelmeanMb} \\
m_{\mathcal{X}_t| x_{t-1}=X_i} &:=&  \int k_\mathcal{X} (\cdot,x_t) p(x_t|x_{t-1}=X_i) dx_t. \label{eq:filtering-conditional-kernel-mean}
\end{eqnarray}
As shown in Example  \ref{ex:conditionalGaussiankernelmean}, since both $k_\mathcal{X}$ and $p(x_t|x_{t-1})$ are Gaussian, the conditional kernel means \eqref{eq:filtering-conditional-kernel-mean} have closed form expressions of the form (\ref{eq:conditionalGaussianAnalyticalkernelmeannaive}). \\[2pt]

\item[{\bf Filtering Step:}]
The task here is to estimate the kernel mean \eqref{eq:filtering-posterior-mean} of the posterior $p(x_t|z_{1:t})$ by applying the KBR (Sect. \ref{sec:KBRpriorprobabilisticmodel}) using (\ref{eq:forwardkernelmeanMb}) as a prior. 
To describe this, define the kernel mean ${m _{{ \mathcal{Z}_{t}}| {{z}_{1:t-1}} }} \in \mathcal{H}_\mathcal{Z}$ of the predictive density $p(z_t|z_{1:t-1}) := \int p(z_t|x_t)p(x_t|z_{1:t-1})dx_{t}$ of a new observation $z_t$:
$$
{m _{{ \mathcal{Z}_{t}}| {{z}_{1:t-1}} }} := \int k_{\mathcal{Z}}(\cdot, z_{t}) p(z_{t} | z_{1:t-1}) dz_{t}.
$$
The KBR first essentially estimates this by applying to the NP-KSR to (\ref{eq:forwardkernelmeanMb}) using the training data $\{ (X_i,Z_i) \}_{i=1}^{n}$; the resulting estimate is
\begin{eqnarray}
{\hat{m} _{{\mathcal{Z}_{t}}| {z_{1:t-1}} }} &:=& \sum_{i=1}^{n} [\boldsymbol{\beta}_{{\mathcal{Z}_{t}}| {{z}_{1:t-1}} }]_i k_{\mathcal{Z}}(\cdot, Z_i), \nonumber \\
 {{\boldsymbol{\beta }}_{{\mathcal{Z}_{t}}| {{z}_{1:t-1}} }} &:=& {\left( {{G_X} + n\varepsilon {I_n}} \right)^{ - 1}}{G_{X'| X} \boldsymbol{\alpha}_{{\mathcal{X}_{t-1}}| {{z}_{1:t-1}} }} \in \mathbb{R}^n, \label{eq:predictive-weight-filtering}
\end{eqnarray}
where $G_X = (k_\mathcal{X}(X_i,X_j))_{i,j=1}^n \in \mathbb{R}^{n \times n}$ and $G_{X'| X} \in \mathbb{R}^{n \times n}$ is defined by evaluations of the conditional kernel means \eqref{eq:filtering-conditional-kernel-mean}: $(G_{X'| X})_{ij} :=  m_{\mathcal{X}_t| x_{t-1}=X_j}(X_i) = \int k_\mathcal{X} (X_i,x_t) p(x_t|x_{t-1}=X_j) dx_t$.
(If $t=1$, generate sample points $\tilde{X}_1,\dots,\tilde{X}_n \in \mathcal{X}$ i.i.d.~from $p(x_1)$, and define $G_{X'| X} \in \mathbb{R}^{n \times n}$ as $(G_{X'| X})_{ij} := k(X_i,\tilde{X}_j)$ and $\boldsymbol{\alpha}_{{\mathcal{X}_{t-1}}| {{z}_{1:t-1}}} := (1/n,\dots,1/n)^\top \in \mathbb{R}^n$.)

Using the weight vector ${{\boldsymbol{\beta }}_{{Z_{t}}| {{z}_{1:t-1}} }}$ given above, the KBR then estimates the posterior kernel mean \eqref{eq:filtering-posterior-mean} as
\begin{equation}\label{eq:KBR-estimate-filtering}
{\hat{m} _{{\mathcal{X}_{t}}| {z_{1:t}} }} := \sum_{i=1}^{n} [\boldsymbol{\alpha}_{{\mathcal{X}_{t}}| {{z}_{1:t}} }]_i k_{\mathcal{X}}(\cdot, X_i), \hspace{3mm}
{{\boldsymbol{\alpha }}_{{\mathcal{X}_{t}}| {{z}_{1:t}} }} := {R_{\mathcal{X| Z}}}({{\boldsymbol{\beta }}_{{\mathcal{Z}_{t}}| {{z}_{1:t-1}} }}){{\bf{k}}_Z}(z_t), 
\end{equation}
where ${{\bf{k}}_Z}(z_t) = (k_\mathcal{Z}(Z_i,z_t))_{i=1}^n \in \mathbb{R}^n$ and ${R_{\mathcal{X| Z}}}({{\boldsymbol{\beta }}_{{\mathcal{Z}_{t}}| {{z}_{1:t-1}} }}) \in \mathbb{R}^{n \times n}$ is 
\begin{equation} \label{eq:KBR-matrix-filtering}
 {R_{\mathcal{X| Z}}}({{\boldsymbol{\beta }}_{{\mathcal{Z}_{t}}| {{z}_{1:t-1}} }}) := D({{\boldsymbol{\beta }}_{{\mathcal{Z}_{t}}| {{z}_{1:t-1}} }}){G_Z} { \left( {{{  ( {D({{\boldsymbol{\beta }}_{{\mathcal{Z}_{t }}| {{z}_{1:t-1}} }} ){G_Z}}  )}^2} + \delta {I_n}}  \right)^{ - 1}} \hspace{-1mm}D({{\boldsymbol{\beta }}_{{\mathcal{Z}_{t}}| {{z}_{1:t-1}} }} ),
\end{equation} 
where $G_Z := (k_\mathcal{Z}(Z_i,Z_j))_{} \in \mathbb{R}^{n \times n}$ and $D({{\boldsymbol{\beta }}_{{Z_{t}}| {{z}_{1:t-1}} }}) \in \mathbb{R}^{n \times n}$ is the diagonal matrix with its diagonal elements being ${{\boldsymbol{\beta }}_{{\mathcal{Z}_{t}}| {{z}_{1:t-1}} }}$.
\end{description}

The proposed filtering algorithm is iterative applications of these prediction and filtering steps, as summarized in Algorithm \ref{alg:KBR-FilterTransitionAnalytic}.
The algorithm results in updating the two weight vectors ${{\boldsymbol{\beta }}_{{\mathcal{Z}_{t}}| {{z}_{1:t-1}} }}, {{\boldsymbol{\alpha }}_{{\mathcal{X}_{t }}| {{z}_{1:t}} }} \in \mathbb{R}^n$. 

In Algorithm \ref{alg:KBR-FilterTransitionAnalytic}, the computation of the matrix $G_{X'|X}$ is inside the for-loop for $t=2,\dots,T$, but one does not need to recompute it if the transition model $p(x_t | x_{t-1})$ is invariant with respect to time $t$.
If the transition model depends on time (e.g., when it involves a control signal), then $G_{X'|X}$ should be recomputed for each time.

\begin{algorithm}[t]
   \caption{The proposed filtering method} 
   \label{alg:KBR-FilterTransitionAnalytic}
\begin{algorithmic}
   \STATE {\bfseries Initial Prior Sampling:} Generate $\tilde{X}_1,\dots,\tilde{X}_n \in \mathcal{X}$ i.i.d.~from $p(x_1)$, and define $G_{X'| X} \in \mathbb{R}^{n \times n}$ as $(G_{X'| X})_{ij} := k(X_i,\tilde{X}_j)$ and $\boldsymbol{\alpha}_{{\mathcal{X}_{0}}| {{z}_{1:0}}} := (1/n,\dots,1/n)^\top \in \mathbb{R}^n$.
   \STATE {\bfseries Observe:} \hspace{0mm} ${z}_{1} \in \mathcal{Z}$.
   \STATE {\bfseries Initial Filtering:} Compute ${{\boldsymbol{\alpha }}_{{\mathcal{X}_1}| {{z}_{1}}}} \in \mathbb{R}^n$ by the KBR  (see Eqs.~\eqref{eq:predictive-weight-filtering}\eqref{eq:KBR-estimate-filtering}\eqref{eq:KBR-matrix-filtering}).
   \FOR{ $t = 2:T$}
   \STATE {\bfseries Compute:} $G_{X'| X} \in \mathbb{R}^{n \times n}$ as $(G_{X'| X})_{ij} := \int k_\mathcal{X} (X_i,x_t) p(x_t|x_{t-1}=X_j) dx_t$.
   \STATE {\bfseries Weight Computation 1:} \hspace{0mm} ${{\boldsymbol{\beta }}_{{\mathcal{Z}_{t}}| {{z}_{1:t-1}} }} = {\left( {{G_X} +  n \varepsilon {I_n}} \right)^{ - 1}}{G_{X'| X} }{{\boldsymbol{\alpha }}_{{\mathcal{X}_{t-1}}| {{z}_{1:t-1}} }} \in \mathbb{R}^n$.
   \STATE {\bfseries Observe:}  \hspace{0mm} ${z}_{t} \in \mathcal{Z}$.
   \STATE {\bfseries Weight Computation 2:} \hspace{0mm} ${{\boldsymbol{\alpha }}_{{\mathcal{X}_{t }}| {{z}_{1:t}} }} = {R_{\mathcal{X| Z}}}({{\boldsymbol{\beta }}_{{\mathcal{Z}_{t}}| {{z}_{1:t-1}} }}){{\bf{k}}_Z}({{z}_{t}}) \in \mathbb{R}^n$ (see Eq.~\eqref{eq:KBR-matrix-filtering}).
   \ENDFOR
\end{algorithmic}
\end{algorithm}

\subsection{How to Use the Outputs of Algorithm \ref{alg:KBR-FilterTransitionAnalytic} } \label{sec:ComputationfromOutput}
The proposed filter (Algorithm \ref{alg:KBR-FilterTransitionAnalytic}) outputs a sequence of kernel mean estimates $  {\hat m _{{\mathcal{X}_{1}}| {{z}_{1:1}} }},  {\hat m _{{\mathcal{X}_{2}}| {{z}_{1:2}} }}, \dots, {\hat m _{{\mathcal{X}_{T}}| {{z}_{1:T}} }} \in \mathcal{H}_\mathcal{X}$ as given in \eqref{eq:KBR-estimate-filtering}, or equivalently a sequence of weight vectors $ \boldsymbol{\alpha}_{{\mathcal{X}_{1}}| {{z}_{1:1}} }, \boldsymbol{\alpha}_{{\mathcal{X}_{2}}| {{z}_{1:2}} } \dots, \boldsymbol{\alpha}_{{\mathcal{X}_{T}}| {{z}_{1:T}} } \in \mathbb{R}^n$. 
We describe below two ways of using these outputs. Note that these are not the only ways: e.g., one can also generate samples from a kernel mean estimate using the {\em kernel herding} algorithm \citep{chen2010super}.
See \citet{KernelMeanEmbeddingofDistributions_review2017} for other possibilities.

\begin{enumerate}
\item [(i)] 
The integral (or the expectation) of a function $f \in \mathcal{H}_\mathcal{X}$ with respect to the posterior $p(x_t | z_{1:t})$ can be estimated as (see Sect.~\ref{sec:kernel-mean-estimator})
\begin{eqnarray*}
\int f(x_t) p(x_t|z_{1:t}) dx_t &=& \langle { m _{{\mathcal{X}_{t }}| {{z}_{1:t}} }}, f  \rangle_{\mathcal{H}_\mathcal{X}} \\
&\approx& \langle {\hat m _{{\mathcal{X}_{t }}| {{z}_{1:t}} }}, f  \rangle_{\mathcal{H}_\mathcal{X}}  = \sum_{i=1}^{n} [\boldsymbol{\alpha}_{{X_{t}}| {{z}_{1:t}} }]_i f(X_i).
\end{eqnarray*}
\item [(ii)] 
 A pseudo-MAP (maximum a posteriori) estimate of the posterior $p(x_t|z_{1:t})$ is obtained by solving the preimage problem (see \citet[Sect.~4.1]{KernelBayes'Rule_BayesianInferencewithPositiveDefiniteKernels}).
\begin{eqnarray}
\hat x_{t} := \arg \mathop {\min }\limits_{x \in \mathcal{X}} \| {{k_{\mathcal{X}}}( \cdot ,x) - \hat{m} _{\mathcal{X}_{t}| z_{1:t}}} \|_{\mathcal{{H_X}}}^2 \label{eq:pointestimationfromkernelmean}.
\end{eqnarray}
If for some $C > 0$ we have $k_{\mathcal{X}}(x,x) = C$ for all $x \in \mathcal{X}$ (e.g., when $k_\mathcal{X}$ is a shift-invariant kernel), (\ref{eq:pointestimationfromkernelmean}) can be rewritten as $\hat x_{t} = \arg \mathop {\max }\limits_{x \in \mathcal{X}} \hat{m} _{\mathcal{X}_{t}| z_{1:t}}(x)$. 
If $k_{\mathcal{X}}$ is a Gaussian kernel $k_R$ (as we employ in this paper), then the following recursive algorithm can be used to solve this optimization problem \citep{Mika99kernelpca}:
\begin{eqnarray}
x^{(s+1)}= \frac{\sum_{i=1}^n X_i [\boldsymbol{\alpha}_{{\mathcal{X}_{t}}| {{z}_{1:t}} }]_i k_R(X_i,x^{(s)}) }{\sum_{i=1}^{n} [\boldsymbol{\alpha}_{{\mathcal{X}_{t}}| {{z}_{1:t}} }]_i k_R(X_i, x^{(s)})} \quad (s = 0,1,2,\dots ).
\label{eq:iterationGaussPointEstimate}
\end{eqnarray}
The initial value $x^{(0)}$ can be selected randomly. 
(Another option may be to set $x^{(0)}$ as a point $X_{i_{\rm max}} \in \{ X_1,\dots,X_n \}$ in the training data that is associated with the maximum weight (i.e., $i_{\rm max} =\arg \mathop {\max }\limits  [\boldsymbol{\alpha}_{{\mathcal{X}_{t}}| {{z}_{1:t}} }]_i $). ) 
Note that the algorithm (\ref{eq:iterationGaussPointEstimate}) is only guaranteed to converge to the local optimum, if the kernel mean estimate $\hat{m}_{\mathcal{X}_{t}| z_{1:t}}(x)$ has multiple modes.
\end{enumerate}
   
\section{Experiments} \label{sec:Experiments}
We report three experimental results showing how the use of the Mb-KSR can be beneficial in kernel Bayesian inference when probabilistic models are available.
In the first experiment (Sect.~\ref{sec:Ground-truthExperiment}), we deal with simple problems where we can exactly evaluate the errors of kernel mean estimators in terms of the RKHS norm; this enables rigorous empirical comparisons between the Mb-KSR, NP-KSR and combined estimators. 
We then report results comparing the proposed filtering method (Algorithm \ref{alg:KBR-FilterTransitionAnalytic}) to existing approaches by applying them to a synthetic state space model (Sect.~\ref{Filteringonstatespacemodels}) and to a real data problem of vision-based robot localization in robotics (Sect.~\ref{sec:RobotLocalization}). 

\subsection{Basic Experiments with Ground-truths}
\label{sec:Ground-truthExperiment}

We first consider the setting described in Sect.~\ref{sec:NP-KSR} and \ref{sec:ModelbasedKernelSumRule(Mb-KSR)} to compare the Mb-KSR and the NP-KSR.
Let $\mathcal{X} = \mathcal{Y} = \mathbb{R}^m$. 
Define a kernel $k_\mathcal{X}$ on $\mathcal{X}$ as a Gaussian kernel $k_{R_\mathcal{X}}$ with covariance matrix $R_{\mathcal{X}} \in \mathbb{R}^{n \times n}$ as defined in \eqref{eq:gaussian-kernel}; similarly, let $k_\mathcal{Y} = k_{R_\mathcal{Y}}$ be a Gaussian kernel on $\mathcal{Y}$ with covariance matrix  $R_{\mathcal{Y}} \in \mathbb{R}^{n \times n}$.

Let $p(y|x)$ be a conditional density on $\mathcal{Y}$ given $x \in \mathcal{X}$, which we we define as an additive linear Gaussian noise model:
$p(y|x) = g(y| Ax,\Sigma)$ for  $x,y \in \mathbb{R}^m$, where $\Sigma \in \mathbb{R}^{m \times m}$ is a covariance matrix and $A \in \mathbb{R}^{m \times m}$.
The input density function $\pi(x)$ on $\mathcal{X}$ is defined as a Gaussian mixture $\pi (x) := \sum_{i = 1}^L {{\xi _i}} {g}(x | {\mu _i},{W_i})$, where $L \in \mathbb{N}$, $\xi_i \geq 0$ are mixture weights such that $\sum_{i=1}^L \xi = 1$, $\mu_i \in \mathbb{R}^m$ are mean vectors and $W_i \in \mathbb{R}^{m \times m}$ are covariance matrices.  
Then the output density $q(y) := \int p(y| x) \pi(x)dx$ is also a Gaussian mixture $q(y) = \sum_{i = 1}^L {{\xi _i}{g}(y | A{\mu _i},{\Sigma} + A{W_i}{A^{\top}})}$. 

The task is to estimate the kernel mean ${m_{Q_{\mathcal{Y}}}} = \int k_\mathcal{Y}(\cdot,x) q(y)dy$ of the output density $q(y)$, which has a closed form expression
\begin{align*}
{m_{Q_{\mathcal{Y}}}}= \sum_{i = 1}^L {{\xi _i}{g}(\cdot | A{\mu _i},{R_{\mathcal{Y}}} + {\Sigma} + A{W_i}{A^\top})}. \label{eq:analyticalkernelmeanlinearGaussian}
\end{align*}
This expression is used to evaluate the error  $\left\| {{m_{Q_{\mathcal{Y}}}} - {{\hat m}_{Q_{\mathcal{Y}}}}} \right\|_{{ \mathcal{H_Y}}}$ in terms of the distance of the RKHS $\mathcal{H}_\mathcal{Y}$, where ${\hat m}_{Q_{\mathcal{Y}}}$ is an estimate given by the Mb-KSR \eqref{eq:MbKSRestimator} or that of the NP-KSR \eqref{eq:KernelSumRuleMapping}.
For the Mb-KSR,  the conditional density $p(y|x)$ is treated as a probabilistic model $p_M(y|x)$, while for the NP-KSR training data are generated for $p(y|x)$; details are explained below.


We performed the following experiment 30 times, independently generating involved data.
Fix parameters $m=2$, $A= \Sigma=I_2$, $L=4$, $\xi_1 = \cdots = \xi_4 = 1/4$,  $R_{\mathcal{X}}=0.1I_2$ and $R_{\mathcal{Y}}=I_2$.
We generated training data $\{(X_i, Y_i)\}_{i=1}^{500}$ for the conditional density $p(y|x)$ by independently sampling from 
the joint density $p(x,y) := p(y|x)p(x)$, where $p(x)$ is the uniform distribution on $[-10,10]^2 \subset \mathcal{X}$. 
The parameters in each component of $\pi (x) = \sum_{i = 1}^L {{\xi _i}} {g}(x | {\mu _i},{W_i})$ were randomly generated as $\mu_i \mathop \sim \limits^{i.i.d.} \mathrm{Uni}[-5,5]^2$ ($i=1,2,3,4$) and $W_i=U^{\top}_{i} U_i$ with $U_i \mathop \sim \limits^{i.i.d.} \mathrm{Uni}[-2,2]^4$ ($i=1,2,3,4$), where ``$\mathrm{Uni}$'' denotes the uniform distribution.
The input kernel mean $m_\Pi := \int k(\cdot,x)\pi(x)dx$ was then approximated as $\hat m_{\Pi} = \frac{1}{500}\sum_{i=1}^{500}k_{\mathcal{X}}(\cdot, \tilde X_i)$, where $\tilde X_1 \ldots \tilde X_{500} \in \mathcal{X}$ were generated independently from $\pi(x)$.


\begin{figure}[t]
\begin{center}
\includegraphics[width =12cm, angle = 0]{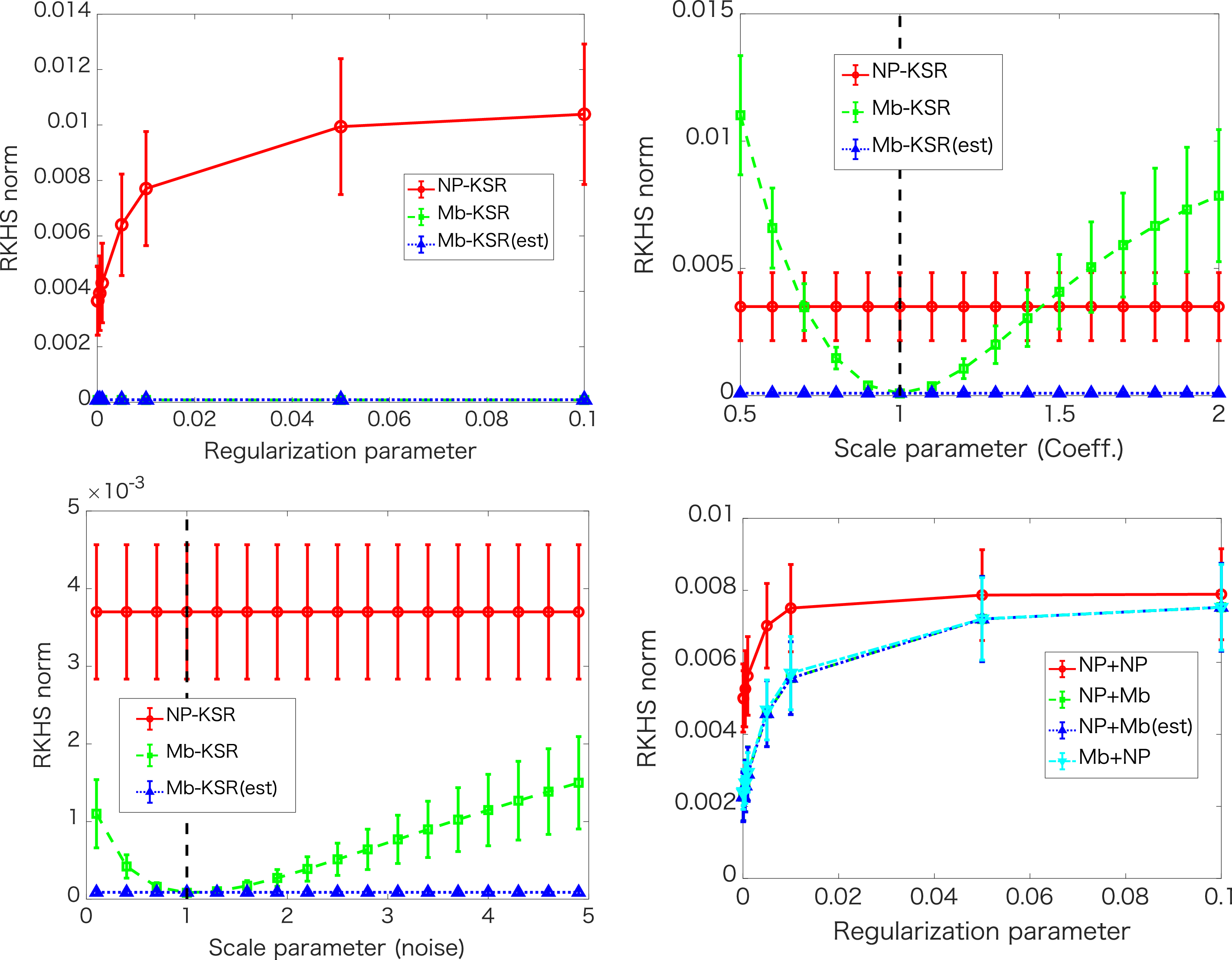}

\caption{Top left: Estimation errors $\left\| {{m_{Q_{\mathcal{Y}}}} - {{\hat m}_{Q_{\mathcal{Y}}}}} \right\|_{{ \mathcal{H_Y}}}$ vs. regularization constants $\epsilon$. The errors of the Mb-KSR and the Mb-KSR (est) are very small and overlap each other.
Top right: A model misspecification case (estimation errors vs. scale parameters $\sigma_1>0$). 
Bottom left: A model misspecification case (estimation errors vs. scale parameters $\sigma_2>0$). 
Bottom right: Estimation errors $\left\| {{m_{Q_{\mathcal{Z}}}} - {{\hat m}_{Q_{\mathcal{Z}}}}} \right\|_{{ \mathcal{H_Z}}}$ vs. regularization constants $\epsilon$ for combined estimators.  
The errors of three estimators (i) NP-KSR and Mb-KSR, (ii) NP-KSR and estimated Mb-KSR and (iii) Mb-KSR and NP-KSR are very close and thus overlap each other.
In all the figures, the error bars indicate the standard deviations over 30 independent trials. }
\label{fig:groundtruth_experiment_4Fig.eps}
\end{center}
\end{figure}

Fig.~\ref{fig:groundtruth_experiment_4Fig.eps} (top left) shows the averages and standard deviations of the error $\left\| {{m_{Q_{\mathcal{Y}}}} - {{\hat m}_{Q_{\mathcal{Y}}}}} \right\|_{{ \mathcal{H_Y}}}$ over the 30 independent trials, with the estimate  ${{\hat m}_{Q_{\mathcal{Y}}}}$ given by three different approaches: NP-KSR, Mb-KSR and ``Mb-KSR (est).''
The NP-KSR learned $p(y|x)$ with using the training data $\{(X_i, Y_i)\}_{i=1}^{500}$, and we report results with different regularization constants as $\varepsilon = [.1,.05,.01,.005,.001, .0005, .0001, .00005]$ (horizontal axis).
For the Mb-KSR, we used the true $p(y|x)$ as a probabilistic model $p_M(y|x)$.
``Mb-KSR (est)'' is the Mb-KSR with $p_M(y|x)$ being the linear Gaussian model with parameters $A$ and $\Sigma$ learnt from $\{(X_i, Y_i)\}_{i=1}^{500}$ by maximum likelihood estimation.

We can make the following observations from Fig.~\ref{fig:groundtruth_experiment_4Fig.eps} (top left): 
1) If the probabilistic model $p_M(y|x)$ is given by parametric learning with a well-specified model, then the performance of the Mb-KSR is as good as that of the Mb-KSR with a correct model;
2) While the NP-KSR is a consistent estimator, its performance is worse than the Mb-KSR, possibly due to the limited sample size and the nonparametric nature of the estimator;
3) The performance of the NP-KSR is sensitive to the choice of a regularization constant.

We next discuss results highlighting the Mb-KSR using misspecified probabilistic models, shown in Fig.~\ref{fig:groundtruth_experiment_4Fig.eps} (top right and bottom left).
Here the NP-KSR used the best regularization constant in Fig.~\ref{fig:groundtruth_experiment_4Fig.eps} (top left), and the Mb-KSR (est) was given in the same way as above.
In Fig. \ref{fig:groundtruth_experiment_4Fig.eps} (top right), the Mb-KSR used a misspecified model defined as
$p_M(y|x) = g(y| \sigma_1 Ax, \Sigma)$, where $\sigma_1 > 0$ controls the degree of misspecification (horizontal axis); $\sigma_1 = 1$ gives the correct model $p(y|x)$ and is emphasized with the vertical line in the figure.
In Fig. \ref{fig:groundtruth_experiment_4Fig.eps} (bottom left), the Mb-KSR used a misspecified model $p_M(y|x) = g(y|  Ax, \sigma_2 \Sigma)$ with $\sigma_2 > 0$; the case $\sigma_2 =1 $ provides the correct model and is indicated by the vertical line.
These two figures show the sensitivity of the Mb-KSR to the model specification, but we also observe that the Mb-KSR outperforms the NP-KSR if the degree of misspecification is not severe.
The figures also imply that, when it is possible, the parameters in a probabilistic model should be learned from data, as indicated by the performance of the Mb-KSR (est).\\[1pt]

\noindent
\textbf{Combined Estimators.}
Finally, we performed experiments on the combined estimators made of the Mb-KSR and NP-KSR described in Sect.~\ref{sec:NP-KSR+Mb-KSR} and \ref{sec:Mb-KSR+NP-KSR}; the setting follows that of these sections, and is defined as follows.

Define the third space as $\mathcal{Z} = \mathbb{R}^m$ with $m = 2$, and let $k_\mathcal{Z} := k_{\mathcal{R}_\mathcal{Z}}$ be the Gaussian kernel \eqref{eq:gaussian-kernel} on $\mathcal{Z}$ with covariance matrix $R_\mathcal{Z} \in \mathbb{R}^{m \times m}$.
Let $p(y|x) := g(y| A_1 x, \Sigma_1 )$ be the conditional density on $\mathcal{Y}$ given $x \in \mathcal{X}$, and $p(z|y) := p(z| A_2 y, \Sigma_2 )$ be that on $\mathcal{Z}$ given $y \in \mathcal{Y}$, both being additive linear Gaussian noise models, where we set $A_1 = A_2 = \Sigma_1 = \Sigma_2 = I_m \in \mathbb{R}^{m \times m}$.
As before, the input density $\pi(x)$ on $\mathcal{X}$ is a Gaussian mixture $\pi (x) = \sum_{i = 1}^L {{\xi _i}} {g}(x | {\mu _i},{W_i})$. Then the output distribution $Q_{\mathcal{Z}}$ is also a Gaussian mixture with $L = 4$ and $\xi_1 = \cdots = \xi_4 = 1/4$, and parameters $\mu_i \in \mathbb{R}^m$ and $W_i \in \mathbb{R}^{m \times m}$ are randomly generated as  $\mu_i \mathop \sim \limits^{i.i.d.} \mathrm{Uni}[-5,5]^2$ and $W_i=U^{\top}_{i} U_i$ with $U_i \mathop \sim \limits^{i.i.d.} \mathrm{Uni}[-2,2]^4$.
Then the output density is given as a Gaussian mixture $q(z) :=\int \int p(z| y) p(y| x)\pi(x) dxdy = \sum_{i = 1}^L {{\xi _i}{g}(z | A_2A_1\mu _i, \Sigma_2+A_2 ({\Sigma_1} + A_1{W_i}{A_1^{\top}})A_2^{\top}})$ .

The task is to estimate the kernel mean $m_{Q_ \mathcal{Z}} := \int k_\mathcal{Z}(\cdot,x)q(z)dz$, whose closed form expression is given as
\begin{align*}
m_{Q_{\mathcal{Z}}} = \sum_{i = 1}^L {\xi _i}{g}  \left (\cdot  |  A_2A_1\mu _i, R_{\mathcal Z} +\Sigma_2+A_2({\Sigma_1} + A_1{W_i}{A_1^{\top}})A_2^{\top}  \right ).     
\end{align*}
The error $\left\| {{m_{Q_{\mathcal{Z}}}} - {{\hat m}_{Q_{\mathcal{Z}}}}} \right\|_{{ \mathcal{H_Z}}}$ as measured by the norm of the RKHS $\mathcal{H}_\mathcal{Z}$ can then also be computed exactly for a given estimate ${{\hat m}_{Q_{\mathcal{Z}}}}$. 

Fig. \ref{fig:groundtruth_experiment_4Fig.eps} (bottom right) shows the averages and standard deviations of the estimation errors over 30 independent trials, computed for four types of combined estimators referred to as ``NP+NP,'' ``NP+Mb,'' ``NP+Mb(est),'' and ``Mb+NP,'' which are respectively (i) NP-KSR + NP-KSR, (ii) NP-KSR + Mb-KSR, (iii) NP-KSR + Mb-KSR (est), and (iv) Mb-KSR + NP-KSR.
As expected, the model-combined estimators (ii)-(iv) outperformed the full-nonparametric case (i).

\subsection{Filtering in a Synthetic State Space Model} \label{Filteringonstatespacemodels}

We performed experiments on filtering in a synthetic nonlinear state space model, comparing the proposed filtering method (Algorithm \ref{alg:KBR-FilterTransitionAnalytic}) in Sect.~\ref{sec:FilteringinStateSpaceModels} with the fully-nonparametric filtering method proposed by \citet{KernelBayes'Rule_BayesianInferencewithPositiveDefiniteKernels}. 
The problem setting, described below,  is based on that of \citet[Sect. 5.3]{KernelBayes'Rule_BayesianInferencewithPositiveDefiniteKernels}.
\begin{itemize}
\item ({\bf State transition process}) 
Let $\mathcal{X} = \mathbb{R}^2$ be the state space, and denote by $x_t:=(u_t, v_t)^\top \in \mathbb{R}^{2}$ the state variable at time $t=1,\dots,T$. 
Let $b, M, \eta, \sigma_h > 0$ be constants.
Assume that each $x_t$ has an latent variable $\theta_t \in [0,2\pi]$, which is an angle.
The current state $x_t$ then changes to the next state $x_{t+1} := ({{u_{t + 1}}}, {{v_{t + 1}}})^{\top}$ according to the following nonlinear model: 
\begin{eqnarray}
({{u_{t + 1}}}, {{v_{t + 1}}})^{\top}\!= \!(1+b\sin(M\theta_{t+1})) ({\cos {\theta _{t + 1}}}, {\sin {\theta _{t + 1}}})^{\top} + {\varsigma _t}, \label{eq:syntheticstatespacemodel}
\end{eqnarray}
where ${\varsigma _t} \sim N({\bf{0}},\sigma_{h}^{2}{I_2})$ is an independent Gaussian noise and
\begin{equation} \label{eq:angle-transition}
{\theta _{t + 1}} ={\theta _t} + \eta \hspace{1mm} (\rm{mod} \hspace{1mm} 2\pi).
\end{equation}

\item ({\bf Observation process}) 
The observation space is $\mathcal{Z} = \mathbb{R}^2$, and let $z_t \in \mathbb{R}^{2}$ be the observation at time $t = 1,\dots,T$. 
Given the current state $x_t:=(u_t, v_t)^\top$, the observation $z_t$ is generated as
\begin{eqnarray*}
z_t =(\mathrm{sign}(u_t)| u_t| ^{\frac{1}{2}},\mathrm{sign}(v_t)| v_t| ^{\frac{1}{2}})^{\top }+ \xi_t,
\end{eqnarray*}
where $\mathrm{sign}(\cdot)$ outputs the sign of its argument, and $\xi_t$ is an independent zero-mean Laplace noise with standard deviation $\sigma_{o} > 0$.  
\end{itemize}
 
We used the fully-nonparametric filtering method by  \citet[Sect. 4.3]{KernelBayes'Rule_BayesianInferencewithPositiveDefiniteKernels} as a baseline, and we refer to it as the {\em fully-nonparametric kernel Bayesian filter (fKBF)}.
As for the proposed filtering method, the fKBR sequentially estimates the posterior kernel means ${m _{{\mathcal{X}_{t}}| {{z}_{1:t}} }} = \int k_{\mathcal{X}}(\cdot, x_{t}) p(x_{t} | z_{1:t}) dx_{t}$ ($t=1,\dots,T$) using the KBR in the filtering step.
The difference from the proposed filter is that the fKBR uses the NP-KSR (Sect.~\ref{sec:NP-KSR}) in the prediction step.
Thus, a comparison between these two methods reveals how the use of a probabilistic model via the Mb-KSR is beneficial in the context of state space models.


We generated training data $(X_{i},Z_{i})_{i=1}^{n} \subset \mathcal{X} \times \mathcal{Z}$ for the observation model as well as those for the transition process $(X_{i},X_{i}')_{i=1}^{n} \subset \mathcal{X} \times \mathcal{X}$ by simulating the above state space model, where $X_{i}'$ denotes the state that is one time ahead of $X_{i}$.
The proposed filter used $(X_{i},Z_{i})_{i=1}^{n}$ in the filtering step, and Eqs.~\eqref{eq:syntheticstatespacemodel} and \eqref{eq:angle-transition} as a probabilistic model in the prediction step.
The fKBF used $(X_{i},Z_{i})_{i=1}^{n}$ in the filtering step, and $(X_{i},X_{i}')_{i=1}^{n}$ in the prediction step.
For each of these two methods, we defined Gaussian kernels $k_{R_\mathcal{X}}$ and $k_{R_\mathcal{Z}}$ of the form \eqref{eq:gaussian-kernel} on $\mathcal{X}$ and $\mathcal{Z}$, respectively, where we set $R_{\mathcal{X}}= \sigma_{\mathcal{X}}^{2}I_{2}$ and $R_{\mathcal{Z}}=\sigma_{\mathcal{Z}}^{2}I_{2}$ for $\sigma_{\mathcal{X}}, \sigma_{\mathcal{Z}} > 0$. 

For each method, after obtaining an estimate $\hat{m}_{\mathcal{X}_t | z_{1:t}}$ of the posterior kernel mean at each time $t = 1,\dots,T$, we computed a pseudo-MAP estimate $\hat x_t$ using the algorithm \eqref{eq:iterationGaussPointEstimate} in Sect.~\ref{sec:ComputationfromOutput}, as a point estimate of the true state $x_t$.
We evaluated the performance of each method by computing the mean squared error (MSE) between such point estimates $\hat x_t$ and true states $x_t$. 
We tuned the hyper parameters in each method (i.e., regularization constants $\delta, \varepsilon > 0$ and kernel parameters $\sigma_{\mathcal{X}}, \sigma_{\mathcal{Y}} > 0$) by two-fold cross validation with grid search. 
We set $T = 100$ for the test phase.

\begin{figure}[t]
\begin{center}
\includegraphics[width =12cm, angle = 0]{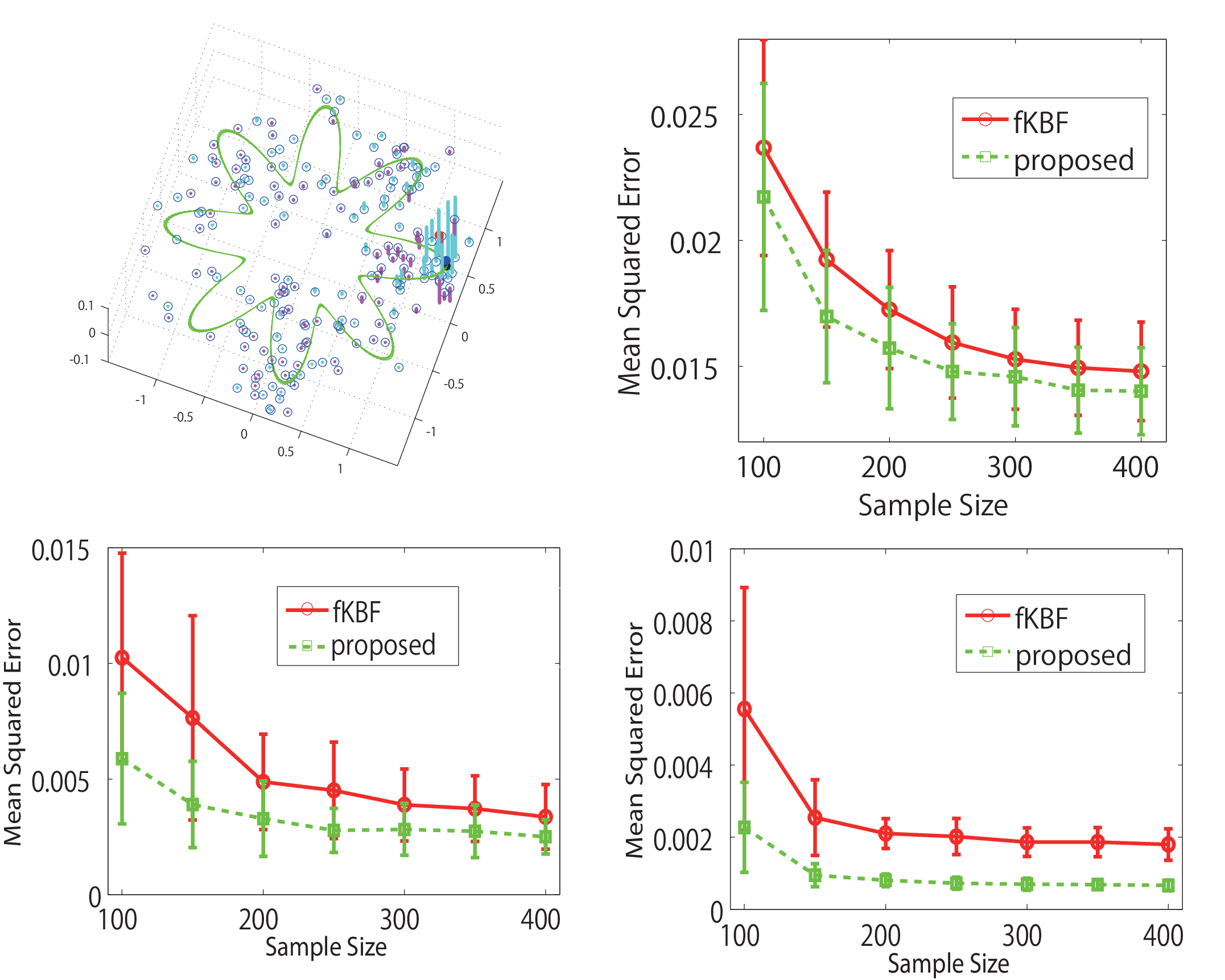}
\caption{Comparisons between the proposed filtering method and the fully-nonparametric kernel Bayes filter (fKBF) by \citet{KernelBayes'Rule_BayesianInferencewithPositiveDefiniteKernels}. For details, see Sect.~\ref{Filteringonstatespacemodels}.
}
\label{fig:Varcircle_TransiGaussMixture001_ObsLaplace001_b=04.eps.eps}
\end{center}
\end{figure}

Fig.~\ref{fig:Varcircle_TransiGaussMixture001_ObsLaplace001_b=04.eps.eps} (top left) visualizes the weight vector $\boldsymbol{\alpha}_{{\mathcal{X}_{t}}| {{z}_{1:t}} } \in \mathbb{R}^n$ of the estimate ${\hat{m} _{{X_{t}}| {z_{1:t}} }} = \sum_{i=1}^{n} [\boldsymbol{\alpha}_{{\mathcal{X}_{t}}| {{z}_{1:t}} }]_i k_{\mathcal{X}}(\cdot, X_i)$ given by the proposed filter \eqref{eq:KBR-estimate-filtering} at a certain time point $t$.
In the figure, the green curve is the trajectory of states given by (\ref{eq:syntheticstatespacemodel}) without the noise term. 
The red and blue points are the observation $z_t$ and the true state $x_t$.
The small points indicate the locations of the training data $X_1,\dots,X_n$, and the value of the weight$ [ \boldsymbol{\alpha}_{{X_{t}}| {{z}_{1:t}} }]_i $ for each data point $X_i$ is plotted in the $z$ axis, where positive and negative weights are colored in cyan and magenta, respectively.


Fig.~\ref{fig:Varcircle_TransiGaussMixture001_ObsLaplace001_b=04.eps.eps} (top right) shows the averages and standard deviations of the MSEs over $30$ independent trials for the two methods, where the parameters of the state space model are $b=0.4$, $M=8$, $\eta=1$, $\sigma_{h}=0.2$ and $\sigma_{o}=0.05$. 
We performed the experiments for different sample sizes $n$.
As expected, the direct use of the transition process \eqref{eq:syntheticstatespacemodel} via the Mb-KSR resulted in better performances of the proposed filter than the fully-nonparametric approach.

Similar results are obtained for Fig.~\ref{fig:Varcircle_TransiGaussMixture001_ObsLaplace001_b=04.eps.eps} (bottom left), where the parameters are set as $b=0.4$, $M=8$, $\eta=1$, $\sigma_{o}=0.01$, and the Gaussian noise ${\varsigma _t}$ in the transition process \eqref{eq:syntheticstatespacemodel} is replaced by a noise from a Gaussian mixture: ${\varsigma _t} \sim \frac{1}{4} \sum_{i=1}^{4}N(\mu_i,(0.3)^{2}{I_2})$ with $\mu_1=(0.2, 0.2)^{\top}$, $\mu_2=(0.2, -0.2)^{\top}$, $\mu_3=(-0.2, 0.2)^{\top}$, and $\mu_4=(-0.2, -0.2)^{\top}$. 
We performed this experiment to show the capability of the Mb-KSR to make use of additive mixture noise models (see Appendix \ref{sec:mixture-noise-models}).   
 
Finally, Fig.~\ref{fig:Varcircle_TransiGaussMixture001_ObsLaplace001_b=04.eps.eps} (bottom right) describes results for the case where we changed the transition model in the test phase from that in the training phase.
That is, we set $b=0.4$, $M=8$, $\sigma_{h}=0.1$, $\sigma_{o}=0.01$ and $\eta=0.1$ in the training phase, but we changed the parameter $\eta$ in \eqref{eq:angle-transition} to $\eta=0.4$ in the test phase.
The proposed filter directly used this knowledge in the test phase by incorporating it by the Mb-KSR, and this resulted in significantly better performances of the proposed filter than the fKBR.
 Note that such additional knowledge in the test phase is often available in practice, for example in problems where the state transition process involves control signals, as for the case of the robot location problem in the next section.
 On the other hand, exploitation of such knowledge is not easy for fully nonparametric approaches like fKBR, since they need to express the knowledge in terms of training samples.

\subsection{Vision-based Robot Localization} \label{sec:RobotLocalization} 
We performed real data experiments on the vision-based robot localization problem in robotics, formulated as filtering in a state space model.
In this problem, we consider a robot moving in a building, and the task is to sequentially estimate the robot's positions in the building in real time, using vision images that the robot has obtained with its camera.

In terms of a state space model, the state $x_t$ at time $t=1,\dots,T$ is the robot's position $x_t :=   ({\tt x}_t, {\tt y}_t,\theta_t) \in \mathcal{X} := \mathbb{R}^2 \times [-\pi,\pi]$, where $({\tt x}_t,{\tt y}_t)$ is the location and $\theta_t$ is the direction of the robot, and the observation $z_t \in \mathcal{Z}$ is the vision image taken by the robot at the position $x_t$. (Here $\mathcal{Z}$ is a space of images.)
It is also assumed the robot records odometry data $u_t := (\bar{\tt x}_t,\bar{\tt y}_t,\bar{\theta}_t) \in \mathbb{R}^2 \times [-\pi,\pi]$, which are the robot's inner representations of its positions obtained from sensors measuring the revolution of the robot's wheels; such odometry data can be used as control signals \citep[Sect.~2.3.2]{ProbabilisticRobotics2005}.
Thus, the robot localization problem is formulated as the task of filtering using the control signals: estimate the position $x_t$ using a history of vision images $z_1,\dots,z_t$ and control signals $u_1,\dots,u_t$ sequentially for every time step $t=1,\dots,T$.

The transition model $p(x_{t+1} | x_t, u_t,u_{t+1})$, which includes the odometry data $u_t$ and $u_{t+1}$ as control signals, deals with robot's movements and thus can be modeled on the basis of mechanical laws; we used an odometry motion model (see e.g.~\citet[Sect.~5.4]{ProbabilisticRobotics2005}) for this experiment, defined as
\begin{eqnarray*}
{\tt x}_{t+1} &=& {\tt x}_t + {\delta _{\mathrm{trans}}}\cos (\theta_t  + {\delta _{\mathrm{rot1}}}) + {\xi _{\tt x}},  \quad {\delta _{\mathrm{rot1}}} := \mathrm{atan} 2(\bar {\tt y}_{t+1} - \bar {\tt y}_t, \bar {\tt x}_{t+1} - \bar {\tt x}_t) - \bar \theta_t, \\
{\tt y}_{t+1} &=&  {\tt y}_t + {\delta _{\mathrm{trans}}}\sin (\theta_t  + {\delta _{\mathrm{rot1}}}) + {\xi _{\tt y}},  \quad {\delta _{\mathrm{trans}}} := ({{{(\bar {\tt x}_{t+1} - \bar {\tt x}_t )}^2} + {{(\bar {\tt y}_{t+1} - \bar {\tt y}_t)}^2}})^{\frac{1}{2}}, \\
\cos \theta_{t+1}  &=& \cos (\theta_t  + {\delta _{\mathrm{rot1}}} + {\delta _{\mathrm{rot2}}}) + {{\xi _c}}, \quad \quad {\delta _{\mathrm{rot2}}}  := \bar \theta_{t+1} - \bar \theta_t  - {\delta _{\mathrm{rot1}}}, \\
\sin \theta_{t+1}  &=& \sin (\theta_t  + {\delta _{\mathrm{rot1}}} + {\delta _{\mathrm{rot2}}}) + {{\xi _s}},
\end{eqnarray*}
where $\mathrm{atan} 2(\cdot, \cdot)$ is the arctangent function with two arguments, and $\xi_{\tt x} \sim N(0,\sigma_{\tt x}^{2})$, ${\xi _{\tt y}}\sim N(0,\sigma_{\tt y}^{2})$ ${\xi _c}\sim N(0,\sigma_c^{2})$, and ${\xi _s}\sim N(0,\sigma_s^{2})$ are independent Gaussian noises with respective variances $\sigma_{\tt x}^{2}, \sigma_{\tt y}^{2}, \sigma_c^{2}$ and $\sigma_s^{2}$, which are the parameters of the transition model.

The observation model $p(z_t|x_t)$ is the conditional probability of a vision image $z_t$ given the robot's position $x_t$; this is difficult to provide a model description in a parametric form, since it is highly dependent on the environment of the building. 
Instead, one can use training data $\{ (X_i, Z_i) \}_{i=1}^n \subset \mathcal{X} \times \mathcal{Z}$ to provide information of the observation model.
Such training data, in general, can be obtained before the test phase, for example by running a robot equipped with expensive sensors or by manually labelling the position $X_i$ for a given image $Z_i$.

\begin{figure}[t]
\begin{center}
\includegraphics[width =7cm, angle = 0]{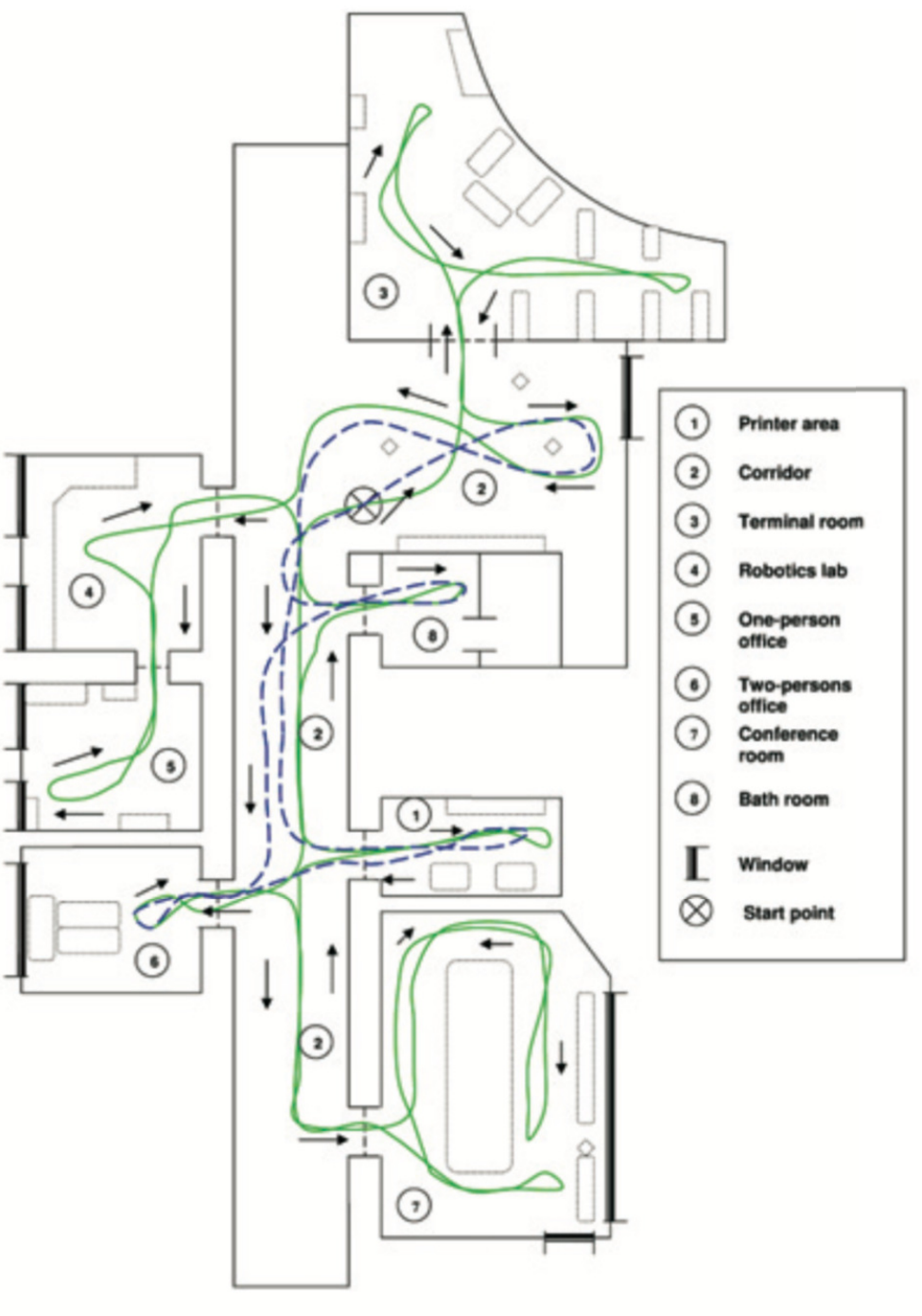}
\caption{Paths that a robot approximately followed for data acquisition \citep[Fig. 1 (b)]{pronobis2009ijrr}. (The use of the figure is granted under the STM Guidelines.)}
\label{fig:map_saarbruecken_a.pdf}
\end{center}

\end{figure}

\begin{figure}[t]
\begin{center}
\includegraphics[width =12cm, angle = 0]{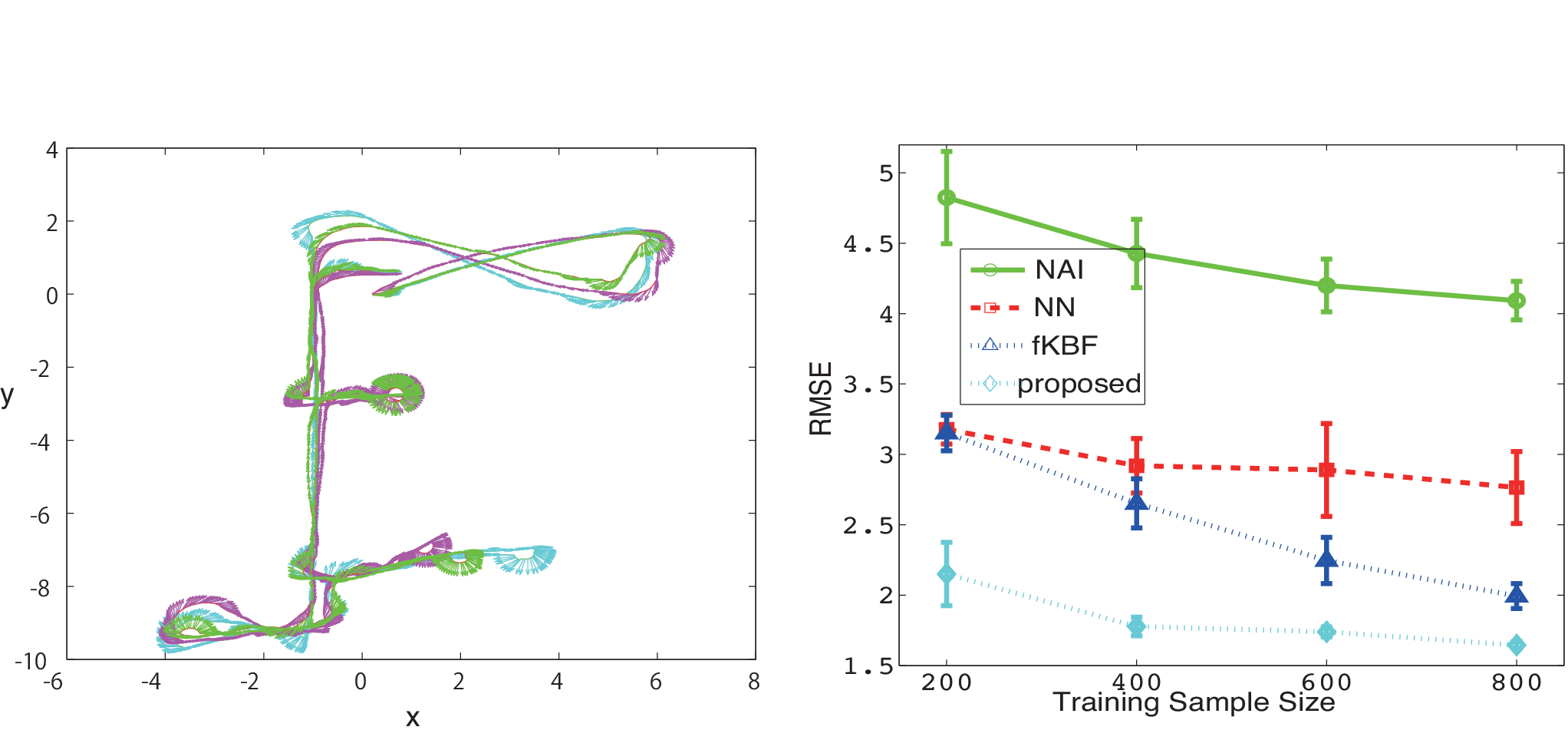}
\caption{(Left) Data for three similar trajectories corresponding to the blue path shown in Fig. \ref{fig:map_saarbruecken_a.pdf}. $(x,y)$ indicates the position of the robot, and the arrow at each position indicates the angle, $\theta$, of the robot's pose. (Right) Estimation accuracy of the robot's position as a function of training sample size $n$.  }
\label{fig:RobotLocalizationResult_Fig2.eps}
\end{center}
\end{figure}

In this experiment we used a publicly available dataset provided by \citet{pronobis2009ijrr} designed for the robot localization problem in an indoor office environment.
In particular, we used a dataset named {\em Saarbr\"ucken, Part A, Standard, and Cloudy}.
This dataset consists of three similar trajectories that approximately follow the blue dashed path in the map described in Fig. \ref{fig:map_saarbruecken_a.pdf}.\footnote{Copyright @ 2009, SAGE Publications.}
The three trajectories of the data are plotted in Fig.~\ref{fig:RobotLocalizationResult_Fig2.eps} (left), where each point represents the robot's position  $({\tt x}_t, {\tt y}_t)$ at a certain time $t$ and the associated arrow the robot's direction $\theta_t$. 
We used two trajectories for training and the rest for testing.


For our method (and for competing methods that use the transition model), we estimated the parameters $\sigma_{\tt  x}^{2}$, $\sigma_{\tt y}^{2}$, $\sigma_c^{2}$ and $\sigma_s^{2}$ in the transition model using the two training trajectories for training by maximum likelihood estimation.
As a kernel $k_\mathcal{Z}$ on the space $\mathcal{Z}$ of images, we used the spatial pyramid matching kernel \citep{Schmid06beyondbags} that is based on the SIFT descriptors \citep{LoweSift2004}, where we set the kernel parameters as those recommended by \citet{Schmid06beyondbags}.
As a kernel $k_\mathcal{X}$ on the space $\mathcal{X}$ of robot's positions, we used a Gaussian kernel.
The bandwidth parameters and regularization constants were tuned by two-fold cross validation using the two training trajectories. 
For point estimation of the position $x_t$ at each time $t=1,\dots,T$ in the test phase, we used the position $X_{i_{\rm max}}$ in the training data $\{ (X_i,Z_i) \}$ associated with the maximum in the weights ${{\boldsymbol{\alpha }}_{{\mathcal{X}_{t}}| {{z}_{1:t}} }}$ for the posterior kernel mean estimate \eqref{eq:KBR-estimate-filtering}:   
$i_{\rm max} = \arg\max_{i = 1,\dots,n} [\boldsymbol{\alpha}_{{\mathcal{X}_{t}}| {{z}_{1:t}} }]_i$


We compared the proposed filter with the following three approaches, for which we also tuned hyper-parameters by cross-validation:
\begin{itemize}
\item \textbf{Na\"ive method (NAI):} 
This is a simple algorithm that estimates the robot's position $x_t$ at each time $t=1,\dots,T$ as the position $X_{i_{\rm max} }$ in the training data that is associated with the image $Z_{i_{\rm max} }$ closest to the given observation $z_t$ in terms of the spatial pyramid matching kernel: $i_{\rm max} := \arg \max_{i=1,\dots,n} k_\mathcal{Z}(z_t, Z_i)$.
This algorithm does not take into account the time-series structure of the problem, was used as a baseline.

\item \textbf{Nearest Neighbors (NN) \citep{Auxiliaryparticlefilterrobotlocalizationfromhigh-dimensionalsensorobservetions}:} 
This method uses the $k$-NN (nearest neighbors) approach to nonparametrically learn the observation model from training data $\{ (X_i,Z_i) \}_{i=1}^n$. 
For the $k$-NN search we also used the spatial pyramid matching kernel.
Filtering is realized by applying a particle filter, using the learned observation model and the transition model (the odometry motion model).
Since the learning of the observation model involves a certain heuristic, this approach may produce biases.

   
\item \textbf{Fully-Nonparametric Kernel Bayes Filter (fKBF) \citep{KernelBayes'Rule_BayesianInferencewithPositiveDefiniteKernels}}: 
For an explanation of this method, see Sect.~\ref{Filteringonstatespacemodels}. 
Since the NP-KSR, which learns the transition model, involves the control signals (i.d., odometry data), we also defined a Gaussian kernel on controls.
As in Sect.~\ref{Filteringonstatespacemodels}, a comparison between this method and the proposed filter reveals the effect of combining the model-based and nonparametric approaches. 
\end{itemize}


Fig.~\ref{fig:RobotLocalizationResult_Fig2.eps} (right) describes averages and standard deviations of RMSEs (root mean squared errors) between estimated and true positions over $10$ trials, performed for different training data sizes $n$.  
The NN outperforms the NAI, as the NAI does not use the time-series structure of the problem. 
The fKBF shows performances superior to the NN in particular for larger training data sizes, possibly due to the fact that the fKBF is a statistically consistent approach. 
The proposed method outperforms the fKBR in particular for smaller training data sizes, showing that the use of the odometry motion model is effective.
The result supports our claim that if a good probabilistic model is available, then one should incorporate it into kernel Bayesian inference.

\section{Conclusions and Future Directions} \label{sec:Summary}

We proposed a method named the model-based kernel sum rule (Mb-KSR) for computing forward probabilities using a probabilistic model in the framework of kernel mean embeddings. 
By combining it with other basic rules such as the nonparametric kernel sum rule and the kernel Bayes rule (KBR), one can develop inference algorithms that incorporate available probabilistic models into nonparametric kernel Bayesian inference. 
We specifically proposed in this paper a novel filtering algorithm for a state space model by combining the Mb-KSR and KBR, focusing on the setting where the transition model is available while the observation model is unknown and only state-observation examples are available. 
We empirically investigated the effectiveness of the proposed approach by numerical experiments that include the vision-based mobile robot localization problem in robotics.

One promising future direction is to investigate applications of the proposed filtering method (or more generally the proposed hybrid approach) in problems where the evolution of states is described by (partial or ordinary) differential equations.
This is a situation common in physical scientific fields where the primal aim is to provide model descriptions for time-evolving phenomena, such as climate science, social science, econometrics and epidemiology.
In such a problem, a discrete-time state space model is obtained by discretization of continuous differential equations, and the transition model $p(x_{t+1}|x_t)$, which is probabilistic, characterizes numerical uncertainties caused by discretization errors.
Importantly, certain numerical solvers of differential equations based on {\em probabilistic numerical methods} \citep{HenOsbGirRSPA2015,CocOatSulGir19,OatSul19} provide the transition model $p(x_{t+1}|x_t)$ in terms of Gaussian probabilities \citep{SchoberDH2014,KerHen16,Schober2018,Tronarp18}. 
Hence, we expect that it is possible to use a transition model obtained from such probabilistic solvers with the Mb-KSR, and to combine a time-series model described by differential equations with nonparametric kernel Bayesian inference.

Another future direction is to extend the proposed filtering method to the {\em smoothing} problem, where the task is to compute the posterior probability over state trajectories, $p(x_1,\dots,x_T | z_1, \dots, z_T)$.
This should be possible by incorporating the the Mb-KSR into the fully-nonparametric filtering method based on kernel Bayesian inference developed by \citet{KernelBayesSmoothingAISTATS2016}.
An important issue related to the smoothing problem is that of estimating the parameters of a probabilistic model in hybrid kernel Bayesian inference.
For instance, in the smoothing problem, one may also be asked to estimate the parameters in the transition model from a given test sequence of observations.
We expect that this can be done by developing an EM-like algorithm, or by using the ABC-based approach to maximum likelihood estimation proposed by \citet{KajYamKanFuk18}.

\appendix


\section{Conditional Kernel Means for Additive Noise Models}
\label{sec:ConditionalKernelMeanofProbabilisticModel} 
\label{sec:Appendix}
While we focused on the additive Gaussian model with a Gaussian kernel in the main body, we collect here other noise models and the corresponding kernels that can be used with the Mb-KSR. 
The key is how to find pairs of a probability density $p$ and a kernel $k$, both of which are defined on $\mathbb{R}^m$, such that the the kernel mean $m_p(x) := \int k(x,y)p(y)dy$ has a closed form expression.
To this end, we briefly mention  \citet{CharacteristicKernelsandInfinitelyDivisibleDistributions}, who study certain such pairs. 

The idea of \citet{CharacteristicKernelsandInfinitelyDivisibleDistributions} is to find pairs of a density $p$ and a shift-invariant kernel $k$ such that both $p$ and $k$ share the same functional form; such pairs are called {\em conjugate}.
Recall that a kernel $k$ is shift-invariant if there exists a function $\kappa:\mathbb{R}^m \to \mathbb{R}$ such that $k(x,y) = \kappa(x-y)$ for $x,y \in \mathbb{R}^m$; see \citet[Section 4.2]{GaussianProcessesforMachineLearning} for examples of such kernels.
In this case the kernel mean $m_p$ can be written as the convolution between $\kappa$ and $p$: $m_p(x) =  (\kappa*p)(x) = \int \kappa(x-y)p(y)dy$.
Therefore one can find pairs of $k$ and $p$ that admit a closed form expression of $m_p$ by examining a convolution semigroup (i.e., a family of density functions that is closed under convolution) in which the function $\kappa$ is included. 
For instance, the set of Gaussian densities is closed under convolution, and therefore the kernel mean $m_p = \kappa*p$ of a Gaussian density $p$ has a closed form expression (which is again Gaussian) if $\kappa$ is also Gaussian.

Examples other than those described below may be found in Table 1 of \citet{Probabilistic_Integration:_A_Role_for_Statisticians_in_Numerical_Analysis}, which collects pairs of a kernel and a density whose kernel means have closed form expressions.

\subsection{Cauchy Noise Models and Rational Quadratic Kernels}
Let $\mu \in \mathbb{R}^m$ and $\Sigma \in \mathbb{R}^{m \times m}$ be a positive definite matrix.
The density function of a Cauchy distribution on $\mathbb{R}^m$ (with $\mu$ and $\Sigma$ being its location and scale parameters) is defined as
\begin{equation} \label{eq:caucahy-density}
p_{\rm Cauchy}(x | \mu, \Sigma) = C_{m,\Sigma} (1 + (x-\mu)^\top \Sigma^{-1} (x-\mu) )^{- \frac{1+m}{2} }, 
\end{equation}
where $C_{m,\Sigma} :=  \frac{\Gamma( (1+m)/ 2 )}{\Gamma(1/2) \pi^{m/2} |\Sigma|^{1/2} }$ is the normalization constant. 
Let $f:\mathbb{R}^m \to \mathbb{R}^m$ be a known function.
Then an additive Cauchy noise model, which is a conditional density function on $\mathcal{Y} = \mathbb{R}^m$ given $x \in \mathcal{X} = \mathbb{R}^m$, is defined as
\begin{equation} \label{eq:additive-cauchy}
p_M(y|x) := p_{\rm Cauchy}( y | f(x), \Sigma ).    
\end{equation}

For a positive definite matrix $R \in \mathbb{R}^{m \times m}$, denote by $k_R: \mathbb{R}^m \times \mathbb{R}^m \to \mathbb{R}$ be a normalized rational quadratic kernel \citep[Eq.~4.19]{GaussianProcessesforMachineLearning} defined as
\begin{equation*} 
    k_{R}(x_{1},x_{2})= p_{\rm Cauchy}(x_{1}-x_{2}| 0,R), \quad x_1,x_2 \in \mathbb{R}^m,
\end{equation*}
where $ p_{\rm Cauchy}$ is the Cauchy density \eqref{eq:caucahy-density}.
This kernel can be written as a scale mixture of Gaussian kernels with different bandwidth parameters; see \citet[p.~87]{GaussianProcessesforMachineLearning}. 
Then, if $R = \gamma^2 \Sigma$, the conditional kernel mean  \eqref{eq:cond-mean-model} with $k_\mathcal{Y} := k_{R}$ is given by
\begin{eqnarray*}
{m _{\mathcal{Y}| x}} (y) = p_{\rm Cauchy}( y
| f(x),(1+\gamma)^2 \Sigma ), \quad x,y \in \mathbb{R}^m.  
\end{eqnarray*}
See \citet[Example 4.3]{CharacteristicKernelsandInfinitelyDivisibleDistributions} for details and for a generalization to $\alpha$-stable distributions.

\subsection{Variance-Gamma Noise Models and Mat\'ern Kernels}

For $\lambda > m/2$, $\alpha > 0$, $\mu \in \mathbb{R}^m$ and a positive definite matrix $\Sigma \in \mathbb{R}^{m \times m}$, define a {\em variance-gamma distribution} on $\mathbb{R}^m$ as
\begin{eqnarray*}
p_{\rm VG}(x | \lambda, \alpha, \mu, \Sigma) &:=& \frac{2^{1-\lambda}}{(2\pi)^{m/2} \Gamma(\lambda)} \alpha^{\lambda + m/2} \left[ ( x - \mu )^\top \Sigma^{-1} (x-\mu) \right]^{ (\lambda - m/2)/2} \\
&& \times K_{\lambda-m/2} \left( \alpha \left[ ( x - \mu )^\top \Sigma^{-1} (x-\mu) \right]^{1/2} \right), \quad x \in \mathbb{R}^m,
\end{eqnarray*}
where $ K_{\lambda-m/2}$ is the modified Bessel function of third kind with index $\lambda - m/2$; this is obtained as a specific case of \citet[Eq.~2.4, p.74]{Hammerstein2010} with the asymmetry parameter $\beta = 0$. 
Note that for $\lambda = (m+1)/2$ and $\alpha = 1$, the variance gamma distribution reduces to a Laplace distribution.

The form of the variance-gamma distributions is the same as that of Mat\'ern kernels \citep{Mat86}. 
In fact, the Mat\'ern kernel described in \citet[Eq.~4.14]{GaussianProcessesforMachineLearning} is, up to constant, given by 
\begin{equation}\label{eq:Matern}
k(x_1,x_2) = p_{\rm VG}(x_1 - x_2 | \nu + m/2, \sqrt{2\nu}, 0, \Sigma), \quad x_1, x_2 \in \mathbb{R}^m,
\end{equation}
where $\nu > 0$; $\nu + m/2$ is the order of differentiability of functions in the associated RKHS (which is norm-equivalent to a Sobolev space).
Note also that the Laplace kernel is the Mat\'ern kernel with $\nu = 1/2$. 

For a known function $f: \mathbb{R}^m \to \mathbb{R}^m$, we define an additive variance-gamma noise model as
$$
p_M(y|x) := p_{\rm VG}( y | \lambda, \sqrt{2\nu}, f(x), \Sigma ).
$$
Then with the Mat\'ern kernel \eqref{eq:Matern}, the conditional kernel mean  \eqref{eq:cond-mean-model} is given by
\begin{eqnarray*}
{m _{\mathcal{Y}| x}} (y) = p_{\rm VG}( y
|  \lambda + \nu+m/2, \sqrt{2\nu}, f(x), \Sigma ), \quad x,y \in \mathbb{R}^m.  
\end{eqnarray*}
See \citet[Example 4.6]{CharacteristicKernelsandInfinitelyDivisibleDistributions} for details.


\subsection{Mixture Noise Models} \label{sec:mixture-noise-models}
For a known function $f: \mathbb{R}^m \to \mathbb{R}^m$, consider a probabilistic model
\begin{equation} \label{eq:mixture-noise-model}
p_M(y|x) = p_{\rm mix}(y-f(x)), \quad x,y \in \mathbb{R}^m,
\end{equation}
where $p_{\rm mix}$ is a mixture density
$$
p_{\rm mix}(y) = \sum_{i=1}^L \omega_i p_i(y), \quad y \in \mathbb{R}^m,
$$
with $\omega_1,\dots,\omega_L \geq 0$ are mixing coefficients such that $\sum_{i=1}^L \omega_i = 1$ and $p_1, \dots, p_L$ are probability density functions on $\mathbb{R}^m$.
For a kernel $k_\mathcal{Y}$ on $\mathbb{R}^m$, the conditional kernel mean of the mixture model \eqref{eq:mixture-noise-model} is then given by 
$$
{m _{\mathcal{Y}| x}} (y) 
= \int k_\mathcal{Y}(\cdot,y) p_{\rm mix}(y-f(x)) dy  = \sum_{i=1}^L \omega_i \int k_\mathcal{Y}(\cdot,y) p_i (y-f(x)) dy.
$$
Therefore, if the terms $\int k_\mathcal{Y}(\cdot,y) p_i (y-f(x)) dy$ admit closed form expressions (e.g., when both $k_\mathcal{Y}$ and $p_1,\dots,p_n$ are Gaussian), then the conditional kernel mean is also given in closed form.

\section{Proof of Proposition \ref{Prop:consistencyMb-KSR}} \label{subsec:consistency}

\begin{proof}
We can expand the squared error in the RKHS $\mathcal{H}_\mathcal{Y}$ as
\begin{eqnarray*}
&& \| \hat{m}_{Q_{\mathcal{Y}}} -  m_{Q_{\mathcal{Y}}} \|_{\mathcal{H}_\mathcal{Y}}^2 \\
 &=& \left\| {\sum\limits_{i = 1}^\ell {{\gamma_i}{m_{\mathcal{Y}| {{\tilde X}_i}}}} } - {m_{{Q_\mathcal{Y}}}} \right\|_{{\mathcal{H_Y}}}^2  \nonumber  \\
 &=& \sum\limits_{i,j = 1}^\ell {{\gamma_i}} {\gamma_j}{\left\langle {{m_{\mathcal{Y}| {{\tilde X}_i}}},{m_{\mathcal{Y}| {{\tilde X}_j}}}} \right\rangle _{{\mathcal{H_Y}}}} - 2\sum\limits_{i = 1}^\ell {{\gamma_i}} {\left\langle {{m_{\mathcal{Y}| {{\tilde X}_i}}},{m_{{Q_\mathcal{Y}}}}} \right\rangle _{{\mathcal{H_Y}}}} + \left\| {{m_{{Q_\mathcal{Y}}}}} \right\|_{{\mathcal{H_Y}}}^2\\
 &=&  \sum\limits_{i,j = 1}^\ell {{\gamma_i}} {\gamma_j} \int \int_{}^{} {{k_\mathcal{Y}}(y,\tilde y) p_M(y|\tilde{X}_i)  p_M(\tilde{y}|\tilde{X}_j) dy d\tilde{y} } \\
 && - 2\sum\limits_{i = 1}^\ell {{\gamma_i}} \int \left( \int \int_{}^{} {k_\mathcal{Y}}(y,\tilde y) p_M(y|\tilde{X}_i)  p_M(\tilde{y}|x)  dy d\tilde{y} \right) \pi(x) dx  \\
 && + \int \int \left( \int \int_{}^{} {k_\mathcal{Y}}(y,\tilde y)  p_M(y|x) p_M(\tilde{y}|\tilde{x}) dy d\tilde{y} \right) \pi(x) \pi(\tilde{x}) dx d\tilde{x}  \\
 &=& \sum\limits_{i,j = 1}^\ell {{\gamma_i}} {\gamma_j}\theta ({{\tilde X}_i},{{\tilde X}_j}) - 2\sum\limits_{i = 1}^\ell {{\gamma_i}} \int_{}^{} {\theta ({{\tilde X}_i},x)\pi(x) dx }  + \int \int_{}^{} \theta (x,\tilde x) \pi(x) \pi(\tilde{x}) dx d\tilde{x}\\
 &=& {\left\langle {({{\hat m}_\Pi } - {m_\Pi }) \otimes ({{\hat m}_\Pi } - {m_\Pi }),\theta } \right\rangle _{{\mathcal{H_X}} \otimes {\mathcal{H_X}}}}\\
 &\le& 
 \left\| {{{\hat m}_\Pi } - {m_\Pi }} \right\|_{{\mathcal{H_X}}}^2\left\| \theta  \right\|_{{\mathcal{H_X}} \otimes {\mathcal{H_X}}}^{},
\end{eqnarray*}
where the fifth equality follows from the assumption that $\theta \in \mathcal{H_X} \otimes \mathcal{H_X}$.  
The assertion then follows from $|| \hat m_{\Pi}-m_{\Pi} ||_{\mathcal{H_X}}=O_p(\ell^{-\alpha})$ and $\left\| \theta \right\|_{{\mathcal{H_X}} \otimes {\mathcal{H_X}}}^{} < \infty$.  
\end{proof}

\begin{acknowledgements}
We would like to thank the anonymous reviewers for their comments that helped us improve the clarity and the quality of the paper.
A part of this work was conducted when YN and MK belonged to the Institute of Statistical Mathematics, Tokyo.
\end{acknowledgements}

\bibliographystyle{spbasic}      
\bibliography{reference}   

\end{document}